\documentclass[lettersize,journal]{IEEEtran}
\usepackage{amsmath,amsfonts}

\usepackage{times}
\usepackage{epsfig}
\usepackage{graphicx}
\usepackage{amsmath}
\usepackage{booktabs}
\usepackage{algorithm}
\usepackage{algorithmic}
\usepackage{bm}
\usepackage{multirow}
\usepackage{enumitem}
\usepackage{amssymb}
\usepackage[switch]{lineno} 
\usepackage{threeparttable}
\usepackage{makecell}
\usepackage[caption=false,font=normalsize,labelfont=sf,textfont=sf]{subfig}
\usepackage{textcomp}
\usepackage{stfloats}
\usepackage{url}
\usepackage{verbatim}
\usepackage{cite}
\usepackage{color}
\usepackage{hyperref}

\begin{document}

\title{Self-Supervised Representation Learning \\ with Spatial-Temporal Consistency for Sign Language Recognition}

\author{Weichao~Zhao, Wengang~Zhou, Hezhen~Hu, Min~Wang,
        and~Houqiang~Li,~\IEEEmembership{Fellow,~IEEE}
\thanks{This work is supported by National Natural Science Foundation of China under Contract U20A20183 \& 62021001, and the Youth Innovation Promotion Association CAS. It was also supported by the GPU cluster built by MCC Lab of Information Science and Technology Institution, USTC, and the Supercomputing Center of the USTC. }
\thanks{Weichao~Zhao, Wengang~Zhou, Hezhen~Hu and Houqiang~Li are with CAS Key Laboratory of Technology in GIPAS, the Department of Electrical Engineering and Information Science, University of Science and Technology of China, Hefei, 230027, China (e-mail: saruka@mail.ustc.edu.cn, alexhu@mail.ustc.edu.cn, zhwg@ustc.edu.cn, lihq@ustc.edu.cn).}
\thanks{Min Wang is with the Institute of Artificial Intelligence, Hefei Comprehensive National Science Center, Hefei, 230027, China. (e-mail: wangmin@iai.ustc.edu.cn).}
\thanks{Corresponding author: Wengang Zhou.}}

\markboth{IEEE TRANSACTIONS ON IMAGE PROCESSING,~Vol.~*, No.~*, June~2024}%
{Shell \MakeLowercase{\textit{et al.}}: A Sample Article Using IEEEtran.cls for IEEE Journals}


\maketitle

\begin{abstract}
Recently, there have been efforts to improve the performance in sign language recognition by designing self-supervised learning methods.
However, these methods capture limited information from sign pose data in a frame-wise learning manner, leading to sub-optimal solutions.
To this end, we propose a simple yet effective self-supervised contrastive learning framework to excavate rich context via spatial-temporal consistency from two distinct perspectives and learn instance discriminative representation for sign language recognition.
On one hand, since the semantics of sign language are expressed by the cooperation of fine-grained hands and coarse-grained trunks, we utilize both granularity information and encode them into latent spaces. 
The consistency between hand and trunk features is constrained to encourage learning consistent representation of instance samples.
On the other hand, inspired by the complementary property of motion and joint modalities, we 
\emph{first} introduce first-order motion information into sign language modeling. 
Additionally, we further bridge the interaction between the embedding spaces of both modalities, facilitating bidirectional knowledge transfer to enhance sign language representation.
Our method is evaluated with extensive experiments on four public benchmarks, and achieves new state-of-the-art performance with a notable margin. The source code are publicly available at \href{https://github.com/sakura2233565548/Self-Supervised-Representation-Learning-with-Spatial-Temporal-Consistency-for-SLR}{https://github.com/sakura/Code}.
\end{abstract}


\section{Introduction}
\label{sec:intro}

Sign language serves as the primary communication tool among deaf people.
It is characterized by its unique grammar and lexicon, thus difficult for hearing people to understand.
To facilitate communication between the deaf and the hearing people, sign language recognition~(SLR) is widely studied.
Isolated SLR aims to recognize the meaning of sign language video at the word level.
It is a challenging task due to complex hand gestures, quick motion, and speed variation.
Although current isolated SLR methods have achieved remarkable progress~\cite{li2020word,li2020transferring,koller2018deep,joze2018ms,tunga2021pose,selvaraj-etal-2022-openhands}, they usually suffer over-fitting due to the scarcity of annotated sign data source.

\begin{figure}[t]
	\centering
	\includegraphics[width=0.9\linewidth]{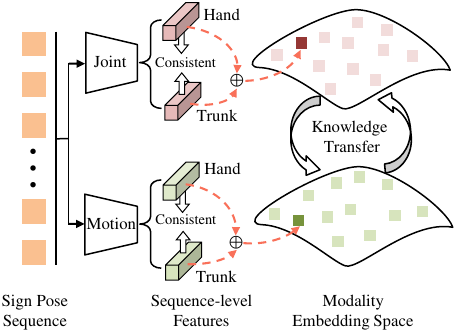}
	\caption{Illustration of our proposed pre-training method. It explicitly mines the spatial-temporal consistency in the sign pose sequence from two perspectives. One focuses on different granularity information from hand and trunk. The other involves different order information from joint and motion modalities.
		Their consistencies are measured in the semantic space for discriminative sign language representation.}
	\label{fig:motivation}
\end{figure}

To mitigate this issue, several works~\cite{albanie2020bsl, hu2021signbert, zhao2023best} have attempted to leverage pre-training techniques to boost the performance on isolated SLR. 
Among them, SignBERT~\cite{hu2021signbert} first explores the self-supervised pre-training in sign language. 
It captures context information through predicting the spatial positions of masked hand joints in a sign pose sequence.
BEST~\cite{zhao2023best} further leverages BERT~\cite{devlin2018bert} pre-training success via utilizing frame-wise discretized pose as the pseudo label.
However, they only mine the contextual information from raw pose spatial locations in a frame-wise manner, which ignores the discriminative spatial-temporal patterns present in sign language video.

In this work, we are dedicated to learning discriminative sign language representation via spatial-temporal consistency modeling 
from two perspectives, as shown in Fig~\ref{fig:motivation}.
On one hand, we collaborate different granularities from both hands and trunk, which preserves the integrity of the sign language meaning.
On the other hand, we bridge different order information from joint and motion modalities. 
Generally, there exist discrepancies in performing lexical signs by different signers in real-world scenarios. As a result, only utilizing static joint modality leads to insufficient representation learning due to the interference of differences in spatial position. The motion information focuses on dynamic movements, which harmoniously complements the joint modality. 

To this end, we propose a self-supervised contrastive learning framework to learn instance discriminative representation from sign pose data.
Specifically, given the sign pose sequence, we first separate both hands and trunk to emphasize these dominant parts. 
Thus, a sign pose sequence is regarded as triplet parts, \textit{i.e.,} right hand, left hand and trunk. 
Meanwhile, we extract the first-order motion information based on the processed triplet joint parts. 
In this way, we construct the sign pose sequence with two modalities, \textit{i.e.,} joint and motion.
Then, we feed them into two branches to separately learn discriminative representation in their embedding space with InfoNCE loss~\cite{oord2018representation}.
Additionally, we constrain the representation of hand and trunk features to guarantee global semantic consistency.
Moreover, considering that joint and motion modalities of a sign pose sequence share the same semantic concept, we transfer available knowledge between the embedding spaces of both modalities to consistently enhance the representative capacity of our framework.

Our contributions are summarized as follows,
\begin{itemize}
	\item We propose a self-supervised pre-training framework. It aims to learn instance discriminative representation via mining spatial-temporal consistency.
	\item We exploit the consistently spatial-temporal information via two main perspectives.
			1) We collaborate fine-grained hand gestures and coarse-grained trunks to represent more holistic sign language meaning. Moreover, both local features are constrained to represent the consistent meaning in a single modality. 
			2) We leverage the complementarity between joint and motion modalities, and further establish representative consistency of both modalities via the knowledge transfer module for comprehensive sign language learning. 
	\item Extensive experiments validate the effectiveness of our proposed method, achieving new state-of-the-art performance on four benchmarks with a notable margin.
\end{itemize}

\section{Related Work}
In this section, we will briefly review several related topics, including sign language recognition and self-supervised representation learning. 

\subsection{Sign Language Recognition}

\textbf{RGB-based methods.} Early works~\cite{starner1995visual,ong2005automatic,rastgoo2021sign} on SLR mainly utilized hand-craft features extracted from hand pose variation and body motion with conventional tracking algorithm, state transition models, \textit{etc}. As the deep convolution neural networks~(CNNs) show their remarkable superiority in computer vision, many works on SLR adopt CNNs as the backbone~\cite{hu2021hand,li2020word,li2020transferring,koller2018deep,li2022transcribing,joze2018ms,sincan2020autsl,bilge2022towards,koller2020quantitative}.
Li~\textit{et al.}~\cite{li2020transferring} utilize a 3D-CNN backbone to extract features with extra knowledge to improve recognition performance.
SEN~\cite{hu2023self} employs a lightweight subnetwork to incorporate local spatial-temporal features into sign language recognition and further improves performance.
Some other methods~\cite{hu2021hand, niu2020stochastic} model the spatial and temporal information separately with 2D CNNs as the backbone. 
NLA-SLR~\cite{zuo2023natural} leverage RGB videos and the human keypoints heatmaps with different temporal receptive fields to build a four-stream framework based on S3D~\cite{xie2018rethinking}, which improves the recognition accuracy through complex model design. Bilge~\textit{et al.,}~\cite{bilge2022towards} propose a zero-shot learning framework on ISLR task, which leverages the descriptive text
and attribute embeddings to transfer knowledge to the instances of unseen sign classes.

\textbf{Skeleton-based methods.}
Pose data is a compact representation of human behavior and contains natural physical connections among skeleton joints~\cite{li2018co,yan2018spatial}.
A bunch of methods~\cite{9975251,9997556,9219176,7450165} have achieved profound impact on skeleton-based action recognition.
Recently, some works~\cite{selvaraj-etal-2022-openhands,hu2021signbert,tunga2021pose,jiang2021skeletor, zhao2023best,9523142} explore the effectiveness of sign pose data for sign language recognition. 
Tunga~\textit{et al.}~\cite{tunga2021pose} utilize GCN and transformer to model spatial-temporal information among sign pose sequences.
HP3D~\cite{lee2023human} extracts expressive 3D human pose to exploit part-specific motion context to enhance sign language representations.
Kindiroglu~\textit{et al.,}~\cite{kindiroglu2024transfer} explores the feasibility of transferring knowledge among different benchmarks with GCN-based approaches. TSSI~\cite{laines2023isolated} converts a skeleton sequence into an RGB image and utilizes an RGB-based backbone to model sequential skeleton features, achieving promising improvement.
Another method, BEST~\cite{zhao2023best}, leverages the BERT pre-training success into SLR and improves recognition accuracy.
SignBERT~\cite{hu2021signbert} and SignBERT+~\cite{hu2023signbert+} propose a generative pretext task for self-supervised pre-training, which reconstructs masked hand pose joints from corrupted pose sequences. 
However, both methods exclusively employ fine-grained hand information within the joint modality to learn frame-wise hand features,  thereby neglecting the incorporation of holistic semantic meaning.

In contrast to them, we leverage different grained and modality information to cultivate a more comprehensive representation. In order to ensure global semantic consistency, we design innovative constraints on inter- and intra-modality. Thus, our proposed model directly learns instance discriminative representation, marking a departure from the limitations posed by the aforementioned methods.

\begin{figure*}[!t]
	\centering
	\includegraphics[width=0.97\linewidth]{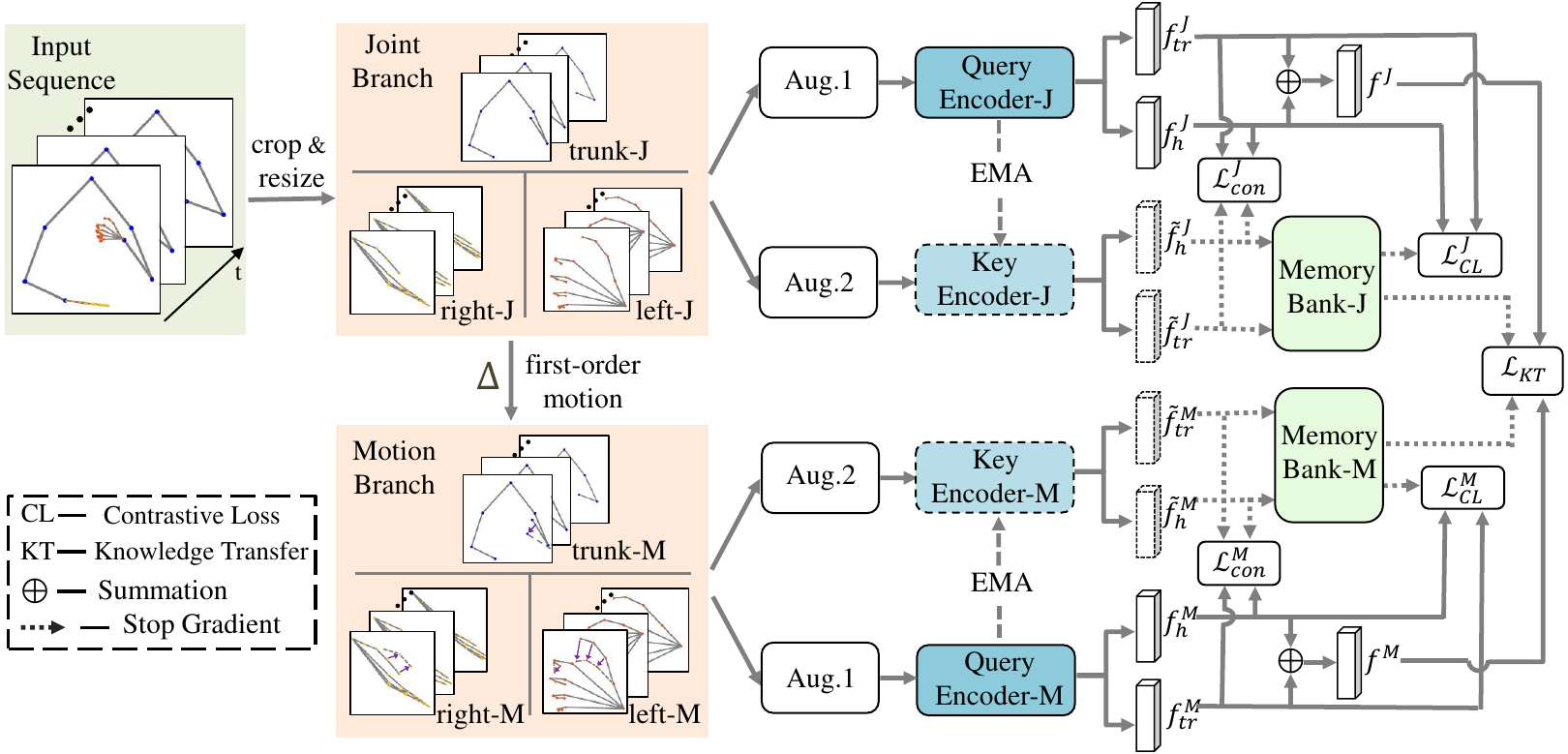}
	\caption{The overall pipeline of the proposed framework during pre-training. The input sequence consists of triplet parts, \textit{i.e.,} both hands and trunk. Meanwhile, we extract the first-order motion from joints in different parts. Then we feed them into two branches to learn instance discriminative representation in a contrastive learning paradigm supervised by contrastive loss $\mathcal{L}_{CL}^J$ and $\mathcal{L}_{CL}^M$, respectively. The key encoder is momentum updated by the query encoder. In addition, we constrain the consistency of hand and trunk features in each branch, \textit{i.e.,} $\mathcal{L}_{con}^J$ and $\mathcal{L}_{con}^M$. Furthermore, we design the bidirectional knowledge transfer module to convey reliable information during cross-modal interaction supervised by $\mathcal{L}_{KT}$. ``J" and ``M" denote the abbreviations of joint and motion.}
	\label{fig:overview}
\end{figure*}

\subsection{Self-supervised Representation Learning}
Self-supervised learning aims to learn feature representation from massive unlabeled data, which usually generates supervision by pretext tasks.
There exist many various pretext tasks for self-supervised representation learning, \textit{i.e.,}  jigsaw puzzles~\cite{noroozi2016unsupervised,noroozi2018boosting}, predicting rotation~\cite{gidaris2018unsupervised}, masked patches reconstruction~\cite{He_2022_CVPR,tong2022videomae,10246363}, contrast similarity between samples~\cite{chen2020simple,zhao2021soda,he2020momentum,oord2018representation, qian2021spatiotemporal,pan2022learning}.
In the video domain, a bunch of advanced works explore how to effectively utilize self-supervised learning techniques for video understanding~\cite{sun2023masked,chen2021rspnet,ding2023prune,zeng2021graph,zeng2019graph,huang2021self,feichtenhofer2021large}.
Among them, MME~\cite{sun2023masked} proposes to reconstruct dense motion trajectories from sparsely sampled videos to capture rich temporal cues to improve the video representation performance.
Moreover, Ding~\textit{et al.}~\cite{ding2023prune} present a strategy to prune spatio-temporal tokens integrally, leading to further computation reduction at an acceptable cost of accuracy degradation.
Recently, contrastive learning methods have attracted increasing attention for video representations due to their simple structure and powerful characterization capability~\cite{qian2021spatiotemporal,li20213d,park2022probabilistic,pan2021videomoco,10100655,10109672}. 
Among them, VideoMoCo~\cite{pan2021videomoco} presents an unsupervised learning method by migrating the success of MoCo~\cite{he2020momentum} from images to videos and achieves competitive performance.
Huang~\textit{et al.}~\cite{huang2021self} decouple motion and contexts in compressed RGB video, to better capture motion information.
Chen~\textit{et al.}~\cite{chen2021rspnet} integrate the relative speed perception task and an appearance-focused task to simultaneously capture motion and appearance cues, enhancing discriminative video representation.
In the sign language domain, current self-supervised learning methods~\cite{jiang2021skeletor, hu2021signbert, albanie2020bsl, zhao2023best} design various masking strategies to learn frame-wise representation, neglecting the importance of dynamic movements. To this end, we draw inspiration from the idea of MoCo~\cite{he2020momentum} and design a novel framework in a contrastive learning paradigm for SLR.

\subsection{Video Consistency Representation Learning}
Video consistency representation learning targets to capture consistent spatial-temporal features from various videos. Generally, there exist two learning paradigms for existing methods: one focuses on understanding the consistency among various transformations in videos, while the other distills knowledge from a pre-trained teacher model. For instance, ASCNet~\cite{huang2021ascnet} proposes to discriminative video features from different playback speeds and appearance information of input videos. HiCo~\cite{qing2022learning} learns a hierarchy of consistencies \textit{i.e.,} visual and topical consistency, in untrimmed videos via constrastive learning. In addition, MVD~\cite{wang2023masked} learns consistent representations of input videos by distilling effective knowledge from pretrained video and image models into the learnable student model.

In our work, the intra-modal constraint loss aims to ensure the global semantic consistency of local features, \textit{i.e.,} hand and trunk poses. The inter-modal knowledge transfer module transfers reliable information from a learnable teacher model, instead of a fixed and well-trained teacher model to each other, thereby learning comprehensive representations of sign pose sequences.

\section{Approach}

\noindent \textbf{Overview.} Fig.~\ref{fig:overview} illustrates our pre-training framework. Given a sign pose sequence, we first crop both hands and trunk parts from the original sequence and resize each part to a fixed size. Meanwhile, we extract the first-order motion sequence based on the processed skeleton sequence. For joint and motion branches, we separately construct two augmented samples of the input sequence through a series of spatio-temporal data augmentations. Next, we feed both augmented samples into the corresponding encoders, i.e., query and key encoder. The outputs of both encoders are utilized for memory-bank contrastive loss to learn instance discriminative representation. In addition, we expect the hand and trunk feature to learn consistent representation with an extra constraint. Furthermore, to enhance sign language representation with joint and motion branches, we design the bidirectional knowledge transfer module to convey useful information between different embedding spaces.

In the following, we will begin by presenting the merits of MoCo~\cite{he2020momentum} as the preliminary work in Sec.~\ref{subsec:preliminary}. Then, we introduce the single modality branch in Sec.~\ref{subsec:modality}. Next, we present the bidirectional reliable knowledge transfer module in Sec.~\ref{subsec:consistency}. Finally, we summarize the overall training objective in Sec.~\ref{subsec:obj} and more model details in Sec.~\ref{subsec:details}.

\subsection{Preliminaries}
\label{subsec:preliminary}
Contrastive learning has been widely utilized thanks to its capability of  instance discrimination. 
As one of the prevailing methods, MoCo~\cite{he2020momentum} achieves impressive performance with several subtle designs.
\romannumeral1)  The momentum updated encoder: given two encoders $Enc\_q(\cdot)$ and $Enc\_k(\cdot)$ that embed $x_q$ and $x_k$ into hidden space: $f = Enc\_q (x_q, \theta_q), \tilde{f} = Enc\_k (x_k, \theta_k)$, in which $\theta$ denotes the learnable parameters of the encoder.  $Enc\_k$ is the updated version of $Enc\_q: \theta_k \leftarrow m \theta_k + (1-m) \theta_q$, where $m$ is a momentum coefficient.
\romannumeral2) The memory bank stores the batch embeddings $\tilde{f}$ in a first-in-first-out~(FIFO) manner in each training step to get rid of redundant computation, and provides negative samples for the next steps.
In this work, we leverage the advantages of MoCo~\cite{he2020momentum} to learn discriminative sign language representation.

\subsection{Single Modality Branch}
\label{subsec:modality}
Given a sign pose sequence $\mathit{V}_{ori}=\{x_i | 1 \leq i \leq T \}$ with the length of $T$ frames, we crop hand and trunk parts to emphasize these dominant regions, then resize them to the fixed resolution. The hand and trunk joint sequences are presented as $\mathit{V}_h^J = \{x_{i,h}|1 \leq i \leq T; h \in \{left, right\}\}$ and $\mathit{V}_{tr}^J = \{x_{i,tr}|1 \leq i \leq T\}$, respectively. Next, the first-order motion information is computed as follows,
\begin{equation}
	\begin{split}
		\mathit{V}_h^M \! \! &= \! \! \{m_{i,h}\! \!:\! x_{i,h} - x_{i-1,h}|2 \leq i \leq T, m_{1,h} = \mathop{\bold{0}}\limits ^{\rightarrow}\}, \\
		\mathit{V}_{tr}^M \! \!  &= \! \! \{m_{i,tr}\! \!:\! x_{i,tr} - x_{i-1,tr}|2 \leq i \leq T, m_{1,tr} = \mathop{\bold{0}}\limits ^{\rightarrow}\},
	\end{split}
\end{equation}

\noindent where $\mathit{V}_h^M$ and $\mathit{V}_{tr}^M$ denote the motion sequences of hand and trunk, respectively. $\mathop{\bold{0}}\limits ^{\rightarrow}$ means that all values of the vector are zero. Finally, four sub-sequences are obtained from the original sign pose sequence, \textit{i.e.,} $\mathit{V}_h^J$, $\mathit{V}_{tr}^J$, $\mathit{V}_h^M$ and $\mathit{V}_{tr}^M$, in which $J$ and $M$ denote the abbreviation of joint and motion, respectively. We feed them into two branches based on the modalities to learn the effective representation in their respective embedding spaces.
Since both branches have a similar structure, for the sake of simplicity, we introduce a single branch of joint modality in the following.

\textbf{Data augmentation.} Given the processed sub-sequences $\mathit{V}_{h}^J$ and $\mathit{V}_{tr}^J$, we utilize the jointly spatio-temporal augmentations to generate two different samples $\mathit{X}_q^J$ and $\mathit{X}_k^J$, each of which includes the augmented hand and trunk sequences. 
\romannumeral1) Spatial augmentations consist of a random combination of rotation, scale, joint mask and random flip operators. It is noted that the random flip simulates the conditions of left- or right-handedness signers. 
\romannumeral2) Temporal augmentations aim to provide dynamic temporal varieties, \textit{i.e.,} sampling frequency and sampling interval, simulating the behavior of different signers. Specifically, we first perform temporal random crop on the input sequence to randomly generate a cropped clip from the interval $[\alpha T, T]$ frames. Then, we sample the fixed length of $T'$ frames by bilinear interpolation among selected clips. 
It is mentioned that augmentations maintain consistent hyper-parameters for the input hand and trunk sub-sequences, while remaining independent for two augmented samples $\mathit{X}_q^J$ and $\mathit{X}_k^J$. 

\textbf{Encoders.} Each branch includes a query encoder $Enc\_q$ and a key encoder $Enc\_k$. Both encoders embed $\mathit{X}_q^J$ and $\mathit{X}_k^J$ into hidden space: $[f^J_{h}, f^J_{tr}] = Enc\_q(\mathit{X}_q^J;\theta^J_q)$ and $[\tilde{\mathit{f}}^J_{h}, \tilde{f}^J_{tr}] = Enc\_k(\mathit{X}_k^J;\theta^J_k)$, where $\mathit{f}^J_{h}, \tilde{\mathit{f}}^J_{h}, \mathit{f}^J_{tr}, \tilde{\mathit{f}}^J_{tr} \in \mathbb{R}^c$.  We adopt ``GCN+Transformer"~\cite{hu2021signbert} architecture with some modifications as an encoder, which is suitable for modeling graph-structure skeleton data. Finally, we obtain the embeddings of the hand and trunk sub-sequences as the output of encoders.

\textbf{Contrastive loss.} Contrastive loss aims to force the model to learn the invariant representation from different data augmentations. It regards one sample's different augments as its positive samples and other samples as negative samples. Since the output of both encoders consists of the emebeddings of hand and trunk, we utilize two memory banks in the joint branch, \textit{i.e.,} $\mathbf{M}_h^J$ and $\mathbf{M}_{tr}^J$. Based on the above settings, the contrastive loss is derived from the InfoNCE loss~\cite{oord2018representation}, which is computed as follows,
\begin{equation}
	\mathcal{L}_{CL}^J \! \! = \! \! \!\! \! \! \sum\limits_{z \in \{\mathit{f}^J_{h}, \mathit{f}^J_{tr}\}} \! \! \! \! \! \! \! - \mathrm{log} \frac{\exp(z^{\mathsf{T}} \tilde{z} / \tau_c)}{\exp(z^{\mathsf{T}} \tilde{z} / \tau_c) + \sum\limits_{i=1}^N(\exp(z^{\mathsf{T}} n_i) / \tau_c)} ,
\end{equation}
where $\tau_c$ is a temperature coefficient~\cite{hinton2015distilling} that controls the distribution of instances and $n_i$ is the key embedding of negative sample stored in the memory bank with respect to $z$. $N$ denotes the size of a queue-based memory bank.

\textbf{Consistency constraint.} Under the supervision of the contrastive loss, the encoder is enforced to learn invariant representation, thereby focusing on semantic information shared among positive pairs. Importantly, complete sign language information relies on the close cooperation of the hand and trunk parts, while simple contrastive supervision only considers the different spatial representations without guaranteeing global semantic consistency. To this end, we design a consistency constraint to keep the similar distribution of both local features in semantic space. Given the local features $\mathit{f}^J_{h}, \mathit{f}^J_{tr}, \tilde{\mathit{f}}^J_h$ and $\tilde{\mathit{f}}^J_{tr}$, we perform this constraint $\mathcal{L}_{con}^J$ by minimizing the KL divergence, which is computed as follows,
\begin{equation}
	\begin{split}
		& \boldsymbol{p}(\mathit{f}, \tau) = \text{softmax}(\mathit{f} / \tau) ; \quad \mathrm{KL}(\boldsymbol{p}||\boldsymbol{q}) = \sum\limits_{i=1}^k p_{i} \cdot \mathrm{log}\frac{p_{i}}{q_{i}}, \\
		& \mathcal{L}_{con}^J \!\!\! =\mathrm{KL}(\boldsymbol{p}(\tilde{\mathit{f}}^J_{h}, \tau_1)||\boldsymbol{p}(\mathit{f}^J_{tr}, \tau_1)) \!+\! \mathrm{KL}(\boldsymbol{p}(\tilde{\mathit{f}}^{J}_{tr}, \tau_1)||\boldsymbol{p}(\mathit{f}^J_{h}, \tau_1)),
	\end{split}
\end{equation}
\noindent where $\text{softmax}(\cdot)$ denotes the function to model the distribution of each feature and $\tau_1$ is a temperature coefficient. 
Moreover, since the $\tilde{\mathit{f}}^J_{h}, \tilde{\mathit{f}}^J_{tr}$ from the key encoder are not trained with gradient, the consistent constraint between local features better guides framework to stable convergence. 

Similar to the joint modality branch, the motion modality branch is trained with the motion data $\mathit{V}_{h}^M$ and $\mathit{V}_{tr}^M$ under the supervision of $\mathcal{L}_{CL}^M$ and $\mathcal{L}_{con}^M$. Though both branches could exploit semantic information from either static joint or dynamic motion modality, the learned representation between both modalities still suffers domain gap, making it difficult to take overall sign language characteristics into account. Moreover, the prominent information from different modalities posses strong complementarity and is beneficial to comprehensive sign language representation. Therefore, it is crucial to facilitate the exchange of effective knowledge stored in the embedding spaces of both modalities.

\subsection{Bidirectional Reliable Knowledge Transfer} 
\label{subsec:consistency}
Joint modality mainly contains the intra-frame spatial position information of the skeleton sequence, while motion modality concentrates more on relative movements of the inter-frame skeleton. If such complementary information, ~\textit{i.e.,} different in joint but similar in motion, and vice versa, could be fully mined and utilized, the holistic sign language representation can be more discriminative and comprehensive. This inspiration enables each modality to transfer reliable knowledge, thereby consistently enhancing available representation.

\textbf{Reliable knowledge modeling.} To convey effective information between modalities, we first need to utilize a proper way to model the knowledge preserved in each modality. Inspired by these knowledge transfer methods~\cite{park2019relational,peng2019correlation,tung2019similarity}, we adopt the pairwise relationship between samples for modality-specific knowledge modeling. For instance, given an embedding $z$ and a set of anchors $\{n_i\}_{i=1}^K$, we compute the similarity between them as $\text{sim}(z, n_i) = z^{\mathsf{T}}n_i$. In our framework, considering that the representation of each modality contains the embeddings of hand and trunk, we regard the summation of both embeddings as the overall meaning of sign pose sequence in each modality  $\mathit{f}^J,\tilde{\mathit{f}}^J,\mathit{f}^M,\tilde{\mathit{f}}^M$, \textit{i.e.,} $\mathit{f}^J \!\! :\mathit{f}^J_h + \mathit{f}^J_{tr}$. Meantime, there are a handful of samples stored in the memory bank. We can easily obtain the required anchors without additional model inference, which is computed as follows,
\begin{equation}
	\begin{split}
		\mathcal{S}^J & = \{s_i^j:x_i + y_i| x_i \in \mathbf{M}_h^J, y_i \in \mathbf{M}_{ tr}^J\}_{i=1}^N, \\
		\mathcal{S}^M & = \{s_i^m:x_i + y_i| x_i \in \mathbf{M}_h^M, y_i \in \mathbf{M}_{ tr}^M\}_{i=1}^N, \\
	\end{split}
\end{equation}  
where $\mathcal{S}^J$ and $\mathcal{S}^M$ denote the set of anchors for joint and motion modality, respectively. In order to ensure that reliable knowledge is established, we select the index of top-$K$ similar embeddings  around a specific embedding $z$ in a set anchor $\mathcal{S}$ as $N_+ = \Gamma(\mathcal{S},  \mathit{z})$.
The resulting pairwise similarities are converted into probability distributions with a temperature coefficient $\tau$, which is computed as follows,
\begin{small}
	\begin{equation}
		\label{equ:5}
		\boldsymbol{p}(z, \tau, \mathcal{S}; N_+) \! \!=\! \! \left\{\! \!\frac{exp(z^{\mathsf{T}}n_i / \tau)}{\sum\nolimits_{j=1}^K exp(z^{\mathsf{T}}n_j / \tau)} \big| n_i,n_j \! \! \in \mathcal{S}, i,j \in N_+\right\},
	\end{equation}
\end{small}

\noindent where $\boldsymbol{p}(z, \tau, \mathcal{S}; N_+)$ describes the distribution around the embedding $z$ in the embedding space of each modality.

\textbf{Knowledge transfer.} Based on the aforementioned probability distribution, we intuitively transfer reliable knowledge by establishing the consistency constraint between both modalities. Different from the previous approaches that transfer knowledge from a fixed and well-trained teacher model to a trainable student, the knowledge in our framework is continuously updated during self-supervised pre-training and each modality can be treated as both teacher and student. 
To this end, we design a bidirectional reliable knowledge transfer module. Specifically, each modality includes two augmented views of the same sign pose sequence encoded into the query $f^J$ or $f^M$, and the key $\tilde{f}^J$ or $\tilde{f}^M$. The key distribution obtained from one modality is utilized to guide the query distribution in another modality, so that the knowledge is transferred accordingly. 

Instancely, for the key embedding $\tilde{f}^A$ from modality A and the query embedding $f^B$ from modality B, we select the top-$K$ nearest neighbors of $\tilde{f}^A$ as anchors and compute the similarity distribution according to Eq.~\ref{equ:5}. The knowledge transfer from modality A to modality B is computed as follows,
\begin{equation}
	\begin{split}
		\! \! & \mathcal{L}_{KT}^{A \rightarrow B} \! \! =  \mathrm{KL}\big(\boldsymbol{p}(\tilde{f}^A \! \! , \tau_t, \mathcal{S}^A; N_+^A)||\boldsymbol{p}(f^B \! \! , \tau_s, \mathcal{S}^B; N_+^A)\big),
	\end{split}
\end{equation}
where $\mathrm{KL}$ denotes the Kullback-Leibler divergence, and $N_+^{A} = \Gamma(\mathcal{S}^A, \tilde{f}^A)$, denotes the index set of top-$K$ anchors around $\tilde{f}^A$. $\tau_t$ and $\tau_s$ are the asymmetric temperatures for teacher and student, respectively. Empirically, we set a smaller temperature for the teacher to emphasize the high-confidence information. Since the knowledge transfer module performs bidirectionally, given the two modalities joint and motion, the bi-reliable knowledge transfer loss is computed as follows,
\begin{equation}
	\mathcal{L}_{KT} = \mathcal{L}_{KT}^{J \rightarrow M} + \mathcal{L}_{KT}^{M \rightarrow J}.
\end{equation}

\subsection{Overall Objective}
\label{subsec:obj}
Overall, the total loss during self-supervised pre-training is the combination of single branches objective and knowledge transfer objective, which is computed as follows,
\begin{equation}
	\mathcal{L} = \underbrace{\lambda_J(\mathcal{L}_{CL}^J + \mathcal{L}_{con}^J)}_{\mathrm{joint}} + \underbrace{\lambda_M(\mathcal{L}_{CL}^M + \mathcal{L}_{con}^M)}_{\mathrm{motion}} + \mathcal{L}_{KT},
\end{equation}
where $\lambda_J$ and $\lambda_M$ are the loss weights, which are set to 0.5 and 0.5, respectively.

\subsection{Model Details}
\label{subsec:details}
During pre-training, an MLP head is attached to each encoder to map the representation into a 128-dimensional embedding. During fine-tuning, we only utilize the query encoder as the pipeline, deprecating the key encoder. We simply replace the final MLP in our framework with a fully-connected layer, and supervise the prediction results of the summation of both branches with the cross entropy objective. 
For clarity, we refer to our method as \textit{Ours}. 
In addition, we utilize a late fusion strategy to enhance the RGB-base methods with the prediction results of our method, namely \textit{Ours}~(+R).
Specifically, we simply sum the prediction results of our method and another RGB-based method.

\section{Experiments}
\subsection{Implementation Details}
\textbf{Data preparation.} Our proposed method utilizes the pose data to represent the hand and body information. Since no available pose annotation is provided in sign language datasets, we utilize the off-the-shelf pose estimator MMPose~\cite{mmpose2020} to extract the 2D pose keypoints. In each frame, the 2D skeleton includes 49 joints, containing 7 trunk joints and 42 hand joints. Moreover, due to the limited regions of both hands in original frame, we crop them based on their coordinates and resize them into a fixed resolution 256$\times$256. Then,we directly compute hand motions according to resized coordinates and normalize them with the fixed resolution. Thus, hand motion mainly contains hand pose variation, the global movement of hand in original frame is contained in wrist joint of trunk part.

\textbf{Parameters setup.}
In our approach, we sample a fixed number of frames $T' = 64$ from each sign language sequence as input. The output dimension of each encoder is 512. The momentum coefficient of key encoder $m$ is set to 0.99. The temperature coefficients $\tau_c$, $\tau_1$, $\tau_t$ and $\tau_s$ are set to 0.07, 0.1, 0.05 and 0.1, respectively. The number of neighbors $K$ and the size of each memory bank are set to 8192 and 16384, respectively. The SGD optimizer~\cite{robbins1951stochastic} is employed with 0.9 momentum. The learning rate is initialized to 0.01 with batch size 64 per GPU, and reduced by a factor of 0.1 every 50 epochs. We totally train 150 epochs in the pre-training stage. 
During fine-tuning, the SGD optimizer with the learning rate of 0.01 and 0.9 momentum is utilized for training. The batch size is set to 64 with total 60 epochs.
All experiments are implemented by PyTorch~\cite{paszke2019pytorch} on NVIDIA RTX 3090. 

\begin{table*}[t!]
	\footnotesize
	\tabcolsep=10.0pt
	\caption{Comparison with state-of-the-art methods on MSASL dataset. ``$\dagger$" indicates the model with self-supervised learning, and ``$\ast$" indicates the method utilized for late fusion. Our method achieves better performance comparison with \\ other self-supervised methods.}
	\begin{center}
		\resizebox{\textwidth}{!}{
			\begin{tabular}{l|cccc|cccc|cccc}
				\toprule
				\multirow{3}{*}[-0.1in]{\makecell[c]{Methods}} & \multicolumn{4}{c|}{MSASL100}
				& \multicolumn{4}{c|}{MSASL200}
				& \multicolumn{4}{c}{MSASL1000}\\ 
				& \multicolumn{2}{c}{P-I} & \multicolumn{2}{c|}{P-C} 
				& \multicolumn{2}{c}{P-I} & \multicolumn{2}{c|}{P-C}
				& \multicolumn{2}{c}{P-I} & \multicolumn{2}{c}{P-C} \\ 
				\cmidrule(lr){2-3} \cmidrule(lr){4-5} \cmidrule(lr){6-7} \cmidrule(lr){8-9} \cmidrule(lr){10-11} \cmidrule(lr){12-13}
				& T-1 & T-5 & T-1 & \multicolumn{1}{c|}{T-5}  
				& T-1 & T-5 & T-1 & \multicolumn{1}{c|}{T-5} 
				& T-1 & T-5 & T-1 & \multicolumn{1}{c}{T-5}    \\ \toprule
				\textbf{Skeleton-based} & & & & & & & & & & & & \\
				ST-GCN~\cite{yan2018spatial} & 59.84 & 82.03 & 60.79 & 82.96  
				& 52.91 & 76.67 & 54.20 & 77.62
				& 36.03 & 59.92 & 32.32 & 57.15 \\ 
				SignBERT~\cite{hu2021signbert}$^\dagger$ & 76.09 & 92.87 & 76.65 & 93.06  
				& 70.64 & 89.55 & 70.92 & 90.00
				& 49.54 & 74.11 & 46.39 & 72.65 \\
				BEST~\cite{zhao2023best}$^\dagger$ & 80.98 & 95.11 & 81.24 & 95.44 & 76.60 & 91.54 &76.75 &91.95 &58.82 &81.18& 54.87 & 80.05\\
				SignBERT+~\cite{hu2023signbert+}$^\dagger$  & 84.94 & 95.77 & 85.23 & 95.76 & 78.51 & 92.49 &79.35 &93.03 &62.42 &83.49& 60.15 & 82.44\\
                    MASA~\cite{zhao2024masa}$^\dagger$ & 83.22 & 95.24 & 83.19 & 95.46 & 79.25 & 92.86 & 79.70 & 93.33 & 63.47 & 83.89 & 60.79 & 83.29 \\
				\textit{Ours} & \textbf{86.26} & \textbf{96.96} & \textbf{86.63} & \textbf{96.79} &\textbf{82.93} &\textbf{94.78} & \textbf{83.43} &\textbf{95.03}  & \textbf{65.22} &\textbf{85.09} & \textbf{62.68} & \textbf{84.38}\\ 
				\midrule
				\textbf{RGB-based} & & & & & & & & & & & &   \\
				I3D~\cite{carreira2017quo}$^\ast$  & - & - & 81.76 & 95.16  
				& - & - & 81.97 & 93.79
				& - & - & 57.69 & 81.05\\ 
				TCK~\cite{li2020transferring}  & 83.04 & 93.46 & 83.91 & 93.52  
				& 80.31 & 91.82 & 81.14 & 92.24
				& - & - & - & - \\ 
				BSL~\cite{albanie2020bsl}  & - & - & - & -  
				& - & - & - & -
				& 64.71 & 85.59 & 61.55 & 84.43 \\
				HMA~\cite{hu2021hand} & 73.45 & 89.70 & 74.59  & 89.70 &66.30  & 84.03 & 67.47 & 84.03 & 49.16 & 69.75  & 46.27 & 68.60 \\
				SignBERT~(+R)~\cite{hu2021signbert}  & 89.56 & 97.36 & 89.96 & 97.51
				& 86.98 & 96.39 & 87.62 & 95.43
				& 71.24 & 89.12 & 67.96 & 88.40 \\
				BEST~(+R)~\cite{zhao2023best} & 89.56 & 96.96 & 90.08 & 97.07 & 86.83 & 95.66 & 87.45 & 95.72 & 71.21 & 88.85 & 68.24 & 87.98 \\
				SignBERT+(+R)~\cite{hu2023signbert+} &90.75 & 97.75 & 91.52 & 97.73 & 88.08 & 96.47 & 88.62 & 96.47 & 73.71 & 90.12 & 70.77 & 89.30 \\
				\textit{Ours}~(+R) & \textbf{90.76} & \textbf{98.41} & \textbf{91.97} & \textbf{98.42} & \textbf{88.82} & \textbf{96.32} & \textbf{89.22} & \textbf{96.37} & \textbf{74.78} & \textbf{90.03} & \textbf{72.18} & \textbf{89.97}\\
				\bottomrule
			\end{tabular}
		}
	\end{center}
	\label{msasl}
\end{table*}

\begin{table}[!h]
	\centering
	\tabcolsep=10pt
	\caption{Statistics of utilized datasets during pre-training. ASL denotes American sign language, and CSL denotes Chinese sign language.}
	\begin{center}
		\resizebox{\linewidth}{!}{
			\begin{tabular}{c|c|c|c|c}
				\toprule
		 	\multicolumn{1}{c|}{Datasets} & \multicolumn{1}{c|}{Train Videos} &         \multicolumn{1}{c|}{Test Videos} & \multicolumn{1}{c}{Language} &
            \multicolumn{1}{c}{Signers}\\
				\midrule
				WLASL~\cite{li2020word} & 21,330 & 4,172 & ASL & 119\\
				MSASL~\cite{joze2018ms}  & 18,205 & 2,878 & ASL & 222 \\
				NMFs-CSL~\cite{hu2021global}  & 25,608 & 6,402 & CSL & 10 \\
				SLR500~\cite{huang2018attention} &  90,000 & 35,000 & CSL & 50 \\ \bottomrule        
			\end{tabular}
                }
	\end{center}
	\label{dataset}
        \vspace{-1.0em}
\end{table}
\subsection{Datasets and Metrics}
\textbf{Datasets.} We conduct experiments on four public sign language benchmarks, \textit{i.e.,} MSASL~\cite{joze2018ms}, WLASL~\cite{li2020word}, NMFs\_CSL~\cite{hu2021global} and SLR500~\cite{huang2018attention}. We utilize the whole training sets of all datasets to pre-train our framework. Tab.~\ref{dataset} presents the overview of the above-mentioned datasets.

MSASL is a large-scale American sign language~(ASL) dataset, consisting of 25,513 samples performed by over 200 signers. The vocabulary includes 1000 words. In particular, it selects the top-$k$ most frequent words with $k=\{100,300\}$, and organizes them as two subsets, namely MSASL100 and MSASL200, respectively. WLASL is another popular ASL dataset with a vocabulary size of 2000. It totally consists of 21,083 samples. Similar to MSASL, it also provides two subsets, named WLASL100 and WLASL300, respectively. Both ASL datasets collect data from websites and bring more challenges due to the unconstrained real-life scenario.

\begin{table*}[h!]
	\footnotesize
	\tabcolsep=9.0pt
	\caption{Comparison with state-of-the-art methods on WLASL dataset. ``$\dagger$" indicates the model with self-supervised learning, and ``$\ast$" indicates the method utilized for late fusion. Our method achieves new state-of-the-art performance on all subsets.}
	\begin{center}
		\resizebox{\textwidth}{!}{
			\begin{tabular}{l|cccc|cccc|cccc}
				\toprule
				\multirow{3}{*}[-0.1in]{\makecell[c]{Methods}} & \multicolumn{4}{c|}{WLASL100}
				& \multicolumn{4}{c|}{WLASL300}
				& \multicolumn{4}{c}{WLASL2000}\\ 
				& \multicolumn{2}{c}{P-I} & \multicolumn{2}{c|}{P-C} 
				& \multicolumn{2}{c}{P-I} & \multicolumn{2}{c|}{P-C}
				& \multicolumn{2}{c}{P-I} & \multicolumn{2}{c}{P-C} \\ 
				\cmidrule(lr){2-3} \cmidrule(lr){4-5} \cmidrule(lr){6-7} \cmidrule(lr){8-9} \cmidrule(lr){10-11} \cmidrule(lr){12-13}
				& T-1 & T-5 & T-1 & \multicolumn{1}{c|}{T-5}  
				& T-1 & T-5 & T-1 & \multicolumn{1}{c|}{T-5} 
				& T-1 & T-5 & T-1 & \multicolumn{1}{c}{T-5}    \\ \toprule
				\textbf{Skeleton-based} & & & & & & & & & & & & \\
				ST-GCN~\cite{yan2018spatial}& 50.78 & 79.07 & 51.62 & 79.47   
				& 44.46 & 73.05 & 45.29 & 73.16
				& 34.40 & 66.57 & 32.53 & 65.45 \\ 
				Pose-TGCN~\cite{li2020word} & 55.43 & 78.68 & - & -   
				& 38.32 & 67.51 & - & -
				& 23.65 & 51.75 & - & -\\ 
				PSLR~\cite{tunga2021pose}& 60.15 & 83.98 & - & -   
				& 42.18 & 71.71 & - & -
				& - & - & - & - \\
				SignBERT~\cite{hu2021signbert}$^\dagger$ & 76.36 & 91.09 & 77.68 & 91.67  
				& 62.72 & 85.18 & 63.43 & 85.71 
				& 39.40 & 73.35 & 36.74 & 72.38  \\ 
				BEST~\cite{zhao2023best}$^\dagger$ & 77.91& 91.47 & 77.83 & 92.50 & 67.66 & 89.22 & 68.31 & 89.57 & 46.25 & 79.33 & 43.52 & 77.65 \\
				SignBERT+~\cite{hu2023signbert+}$^\dagger$  & 79.84 & 92.64 & 80.72 & 93.08 & 73.20 & 90.42 & 73.77 & 90.58 & 48.85 & 82.48 & 46.37 & 81.33\\
				HP3D~\cite{lee2023human} & 76.71 & 91.97 & 78.27 & 92.97 & 67.18 & 89.01 & 67.62 & 89.24 & 44.47 & 79.97 & 42.18 & 78.52\\
                    MASA~\cite{zhao2024masa}$^\dagger$ & 83.72 & 93.80 & 84.47 & 94.30 & 73.65 & 91.77 & 74.33 & 92.13 & 49.06 & 82.90 & 46.91 & 81.80 \\
				\textit{Ours} & \textbf{86.43} & \textbf{95.74} & \textbf{87.13} & \textbf{96.17} & \textbf{77.84} & \textbf{93.86} & \textbf{78.42} & \textbf{94.08} & \textbf{51.98} & \textbf{85.44} & \textbf{49.46} & \textbf{84.32} \\
				\midrule
				\textbf{RGB-based} & & & & & & & & & & & & \\
				I3D~\cite{carreira2017quo}$^\ast$ & 65.89 & 84.11 & 67.01 & 84.58  
				& 56.14 & 79.94 & 56.24 & 78.38
				& 32.48 & 57.31 & - & -\\ 
				TCK~\cite{li2020transferring} & 77.52 & 91.08 & 77.55 & 91.42   
				& 68.56 & 89.52 & 68.75 & 89.41
				& - & - & - & - \\ 
				BSL~\cite{albanie2020bsl} & - & - & - & -  
				& - & - & - & -
				& 46.82 & 79.36 & 44.72 & 78.47\\
				HMA~\cite{hu2021hand} & - & - & - & -
				& - & - & - & - 
				& 37.91 & 71.26 & 35.90 & 70.00 \\
				SignBERT~(+R)~\cite{hu2021signbert}   & 82.56 & 94.96 & 83.30 & 95.00
				& 74.40 & 91.32 & 75.27 & 91.72
				& 54.69 & 87.49 & 52.08 & 86.93 \\
				BEST~(+R)~\cite{zhao2023best} & 81.01 & 94.19 & 81.63 & 94.67 & 75.60 & 92.81 & 76.12 & 93.07 & 54.59 & 88.08 & 52.12 & 87.28 \\ 
				SignBERT+(+R)~\cite{hu2023signbert+} & 84.11 & 96.51 & 85.05 & 96.83 & 78.44 & 94.31 & 79.12 & 94.43 & 55.59 & 89.37 & 53.33 & 88.82 \\
				\textit{Ours}~(+R) & \textbf{88.37} & \textbf{95.74} & \textbf{88.72} & \textbf{96.17}  & \textbf{82.19} & \textbf{94.61}& \textbf{82.64} & \textbf{94.82}  & \textbf{58.06} & \textbf{89.96}& \textbf{55.66} & \textbf{89.62} \\
				\bottomrule 
			\end{tabular}
		}
	\end{center}
	\label{wlasl}
        \vspace{-1.0em}
\end{table*}

\begin{table*}[!t]
	\footnotesize
	\tabcolsep=15pt
	\caption{Comparison with state-of-the-art methods on NMFs-CSL dataset. ``$\dagger$" indicates the model with self-supervised learning, and ``$\ast$" indicates the method utilized for late fusion. Our proposed method shows impressive performance among all methods.}
	\begin{center}
		\resizebox{\textwidth}{!}{
			\begin{tabular}{l|ccc|ccc|ccc}
				\toprule
				\multirow{2}{*}[-0.13in]{Method}      &  \multicolumn{3}{c}{Total} &  \multicolumn{3}{c}{Confusing} &  \multicolumn{3}{c}{Normal}  \\
				\cmidrule(lr){2-4} \cmidrule(lr){5-7} \cmidrule(lr){8-10}
				& T-1 & T-2  & T-5 & T-1 & T-2 &T-5 & T-1 & T-2 & T-5\\ \toprule
				\textbf{Skeleton-based}  & & & & & & & & & \\
				ST-GCN~\cite{yan2018spatial} & 59.9 & 74.7 & 86.8 & 42.2 & 62.3 & 79.4 & 83.4 & 91.3 & 96.7 \\
				SignBERT~\cite{hu2021signbert}$^\dagger$ & 67.0 & 86.8 & 95.3 & 46.4 & 78.2 & 92.1 & 94.5 & 98.1 & 99.6 \\
				BEST~\cite{zhao2023best}$^\dagger$ & 68.5 & - & 94.4 & 49.0 &   - & 90.3 & 94.6 & -  &  99.7 \\ 
                MASA~\cite{zhao2024masa}$^\dagger$ & 71.7  & 89.0 & 97.0 & 53.5 & 81.8  &  92.6 & 95.9 & 98.6 & 99.9 \\
				\textit{Ours} & \textbf{75.6} &\textbf{92.1} & \textbf{97.9} & \textbf{59.4} & \textbf{86.7} & \textbf{96.4} & \textbf{97.3} & \textbf{99.2} &  \textbf{99.6} \\
				\midrule
				\textbf{RGB-based}  & & & & & & & & & \\
				3D-R50~\cite{qiu2017learning}$^\ast$  & 62.1 & 73.2 & 82.9 & 43.1 & 57.9 & 72.4 & 87.4 & 93.4 & 97.0  \\
				I3D~\cite{carreira2017quo}     & 64.4 & 77.9 & 88.0 & 47.3 & 65.7 & 81.8 & 87.1 & 94.3 & 97.3 \\
				TSM~\cite{lin2019tsm}   & 64.5 & 79.5 & 88.7 & 42.9 & 66.0 & 81.0 & 93.3 & 97.5 & 99.0 \\
				GLE-Net~\cite{hu2021global}  & 69.0 & 79.9  & 88.1 & 50.6 & 66.7 & 79.6 & 93.6 & 97.6 & 99.3 \\
				HMA~\cite{hu2021hand} & 64.7 & 81.8 & 91.0 & 42.3 & 69.4 & 84.8 & 94.6 & 98.4 & 99.3 \\
				SignBERT~(+R)~\cite{hu2021signbert}  & 78.4 & 92.0 & 97.3 & 64.3 & 86.5 & 95.4 & 97.4 & 99.3 & 99.9 \\
				BEST~(+R)~\cite{zhao2023best} & 79.2 & -  & 97.1  &  65.5 & - & 95.0 & 97.5  &- & 99.9 \\
				\textit{Ours}~(+R) & \textbf{81.0} & \textbf{93.6} & \textbf{98.0} & \textbf{68.3} & \textbf{89.0} & \textbf{96.5} & \textbf{98.1} & \textbf{99.7} & \textbf{99.9} \\
				\bottomrule
			\end{tabular}
		}
	\end{center}
	\label{NMFs-CSL}
\end{table*}

NMFs\_CSL is a large-scale Chinese sign language~(CSL) dataset with a vocabulary size of 1067 words. All samples are split into 25,608 and 6,402 samples for training and testing, respectively. SLR500 is another largest CSL dataset containing 500 normal words performed by 50 signers. It contains a total of 125,000 samples, of which 90,000 and 35,000 samples are utilized for training and testing, respectively. Different from WLASL and MSASL, both datasets are collected samples from the controlled lab scene.

\textbf{Metrics.} For evaluation, we report the classification accuracy, including Top-1 and Top-5 for downstream SLR task. Specifically, we adopt both per-instance~(P-I) and per-class~(P-C) accuracy metrics following~\cite{li2020transferring,joze2018ms}. Per-instance accuracy is computed over the whole test data, while per-class accuracy is the average of the sign categories present in the test data. For MSASL and WLASL, we report both metrics due to the unbalanced samples for each class. For NMFs\_CSL and SLR500, we report per-instance metric with an equal number of samples per class.

\subsection{Comparison with State-of-the-art Methods} 
In this section, we compare our method with previous state-of-the-art methods on four public benchmarks, including skeleton-based and RGB-based methods..

\textbf{MSASL.} As shown in Tab.~\ref{msasl}, the RGB-based methods BSL~\cite{albanie2020bsl} and TCK~\cite{li2020transferring} achieves better performance than skeleton-based methods, \textit{i.e.,} ST-GCN~\cite{yan2018spatial} and SignBERT~\cite{hu2021signbert}. Despite BEST~\cite{zhao2023best} narrowing the performance discrepancy by utilizing partially available information, there still exists a margin between RGB and skeleton. Compared with BEST, SignBERT+~\cite{hu2023signbert+} also utilizes upper body and hand poses as input, achieving better performance with a larger scale sign pose data during pre-training.
Different from designing various masked strategies, our method learns discriminative representations of sign pose sequence by contrastive learning, which outperforms them with 1.32\%, 4.42\% and 2.80\% Top-1 per-instance accuracy improvement on different subsets, respectively and achieves new state-of-the-art results.

\textbf{WLASL.} 
As shown in Tab.~\ref{wlasl}, ST-GCN~\cite{yan2018spatial} and Pose-TGCN~\cite{li2020word} show poor performance caused by the insufficient learning of sign pose data. TCK~\cite{li2020transferring} and BSL~\cite{albanie2020bsl} transfer knowledge of extra RGB data to enhance the robustness of the model and improve performance. Although SignBERT+~\cite{hu2023signbert+} achieves plausible results, our proposed method outperforms it with 6.59\%, 4.64\% and 3.13\% accuracy increment under the Top-1 per-instance metric. Moreover, \textit{Ours}~(+R) also manifestly achieves the best performance, with 4.26\%, 3.75\% and 2.47\% improvement in WLASL100, WLASL300 and WLASL2000, respectively.

\textbf{NMFs\_CSL.} As shown in Tab.~\ref{NMFs-CSL}, we compare with previous methods under different subsets, in which ``Normal" denotes easily distinguishable sign words, while ``Confusing" denotes more challenging words. GLE-Net~\cite{hu2021global} enhances the discriminative clues from global and local views. SignBERT~\cite{hu2021signbert} and BEST~\cite{zhao2023best} design different masking strategies as pretext tasks with the multi-source sign data to obtain promising performance. Compared with them, our method achieves better performance with 7.1\%, 10.4\% and 2.7\% Top-1 accuracy improvement and new SOTA results.

\textbf{SLR500.} In Tab.~\ref{slr500}, the deep learning methods~\cite{hu2021signbert,zhao2023best,qiu2017learning,hu2021hand} all show comparable results compared with the methods based on hand-crafted features~\cite{laptev2005space,tang2015real}. Among them, the self-supervised methods, \textit{i.e.,} SignBERT~\cite{hu2021global}, SignBERT+~\cite{hu2023signbert+} and BEST~\cite{zhao2023best}, improve the performance by learning latent contextual cues with different masked strategies. Compared with these self-supervised methods, \textit{i.e.,} SignBERT~\cite{hu2021global}, SignBERT+~\cite{hu2023signbert+} and BEST~\cite{zhao2023best}, our method still achieves impressive performance among skeleton-based methods, achieving 96.9\% Top-1 accuracy under the skeleton-based setting.

\textbf{More Discussion.} It is observed that the improvement of Ours~(+R) method is only 0.01\% on MSASL100 compared with SignBERT~(+R)~\cite{hu2023signbert+}. We believe that this phenomenon is caused by two reasons. 1) In MSASL100 and MSASL200 datasets, the accuracy of fusion methods, \textit{i.e.,} SignBERT+~(+R)~\cite{hu2023signbert+} and BEST~(+R)~\cite{zhao2023best} have reached close to 90\%, approaching saturation with limited room for performance growth. Compared to this, Ours~(+R) shows a performance advantage on some more challenging datasets, \textit{i.e.,} MSASL1000, WLASL100, WLASL2000, \textit{etc}. 2) In the late fusion strategy, we employ the same pre-trained I3D model to ensure a fair comparison with previous methods. Consequently, the performance of Ours~(+R) is somewhat constrained by the bottleneck in the capabilities of RGB-based methods.

\begin{table*}[!t]
	\centering
	\small
	\tabcolsep=4pt
	\caption{The performance of different pre-training methods under the linear evaluation protocol on four benchmarks and their subsets. ``linear" denote the evaluation protocol. The ``percent" denotes the proportion of labeled training data utilized during the fine-tuning stage.}
	\begin{center}
		\resizebox{\linewidth}{!}{
				\begin{tabular}{c|c|c|cc|cc|cc|cc|cc|cc|c|c}
					\toprule
					\multirow{2}{*}[-0.05in]{\makecell[c]{Protocol}} & \multirow{2}{*}[-0.05in]{\makecell[c]{\makecell[c]{Percent}}} &
					\multirow{2}{*}[-0.05in]{\makecell[c]{\makecell[c]{Methods}}} & \multicolumn{2}{c|}{MSASL100} & \multicolumn{2}{c|}{MSASL200} & \multicolumn{2}{c|}{MSASL1000} & \multicolumn{2}{c|}{WLASL100} & \multicolumn{2}{c|}{WLASL300} & \multicolumn{2}{c|}{WLASL2000} & \multirow{2}{*}[-0.05in]{\makecell[c]{NMFs-CSL}} & \multirow{2}{*}[-0.05in]{\makecell[c]{SLR500}}\\
					\cmidrule(lr){4-5} \cmidrule(lr){6-7} \cmidrule(lr){8-9} \cmidrule(lr){10-11} \cmidrule(lr){12-13} \cmidrule(lr){14-15}
					&   &  & P-I & P-C &  P-I &  P-C &  P-I &  P-C &  P-I &  P-C &  P-I &  P-C &  P-I &  P-C &  &\\ \midrule
					
					\multirow{4}{*}[-0.1in]{\makecell[c]{linear}} &\multirow{4}{*}[-0.1in]{ $100 \%$} & \multirow{1}{*}[-0.00in]{SignBERT~\cite{hu2021signbert}} & 58.19 & 58.32 & 49.87 & 50.27 & 31.54 & 28.76 & 52.38 & 53.17 & 38.64 & 38.89 & 18.24 & 18.17 & 52.59 & 69.89 \\
					\cmidrule(lr){3-3}
					& & \multirow{1}{*}[-0.00in]{SignBERT+~\cite{hu2023signbert+}} & 66.84 & 66.36 & 59.16 & 59.89 & 34.44 & 32.81 & 57.36 & 57.93 & 45.66 & 46.28 & 24.50 & 24.35 & -- & 71.65 \\
					\cmidrule(lr){3-3}
					& & \multirow{1}{*}[-0.00in]{BEST~\cite{zhao2023best}} & 60.89 & 61.24 & 54.67 & 55.28 & 30.89 & 28.54 & 54.65 & 55.23 & 41.86 & 42.47 & 21.19 & 21.38 & 53.67 & 70.84 \\
					\cmidrule(lr){3-3}
					& & \multirow{1}{*}[-0.00in]{Ours} & \textbf{77.15} & \textbf{76.99}  & \textbf{71.16} & \textbf{71.81} & \textbf{48.30} & \textbf{43.96} & \textbf{72.09} & \textbf{73.27} & \textbf{59.58} & \textbf{60.05} & \textbf{32.35} & \textbf{32.49} & \textbf{71.3} & \textbf{94.3} \\
					\bottomrule     
				\end{tabular}
			}
	\end{center}
	\label{fig:linear}
\end{table*}

\begin{table}[!t]
	\centering
	\footnotesize
	\tabcolsep=20pt
	\caption{Comparison with state-of-the-art methods on SLR500 dataset. ``$\dagger$" indicates the model with self-supervised learning, and ``$\ast$" indicates the method utilized for late fusion.}
	\resizebox{0.8\linewidth}{!}{
		\begin{tabular}{lc}
			\toprule
			Method  &  Accuracy   \\  \toprule
			\textbf{Skeleton-based} & \\
			ST-GCN~\cite{yan2018spatial} &  90.0 \\ 
			SignBERT~\cite{hu2021signbert}$^\dagger$     & 94.5     \\ 
			BEST~\cite{zhao2023best}$^\dagger$  &  95.4   \\ 
			SignBERT+~\cite{hu2023signbert+}$^\dagger$ & 95.4 \\
                MASA~\cite{zhao2024masa}$^\dagger$ & 96.3 \\
			\textit{Ours} & \textbf{96.9} \\
			\textbf{RGB-based}  & \\
			STIP~\cite{laptev2005space}   &  61.8 \\
			GMM-HMM~\cite{tang2015real} &  56.3 \\
			3D-R50~\cite{qiu2017learning}$^\ast$ &  95.1 \\
			HMA~\cite{hu2021hand} & 95.9 \\
			GLE-Net~\cite{hu2021global}   & 96.8      \\
			SignBERT~(+R)~\cite{hu2021signbert} & 97.6 \\
			BEST~(+R)~\cite{zhao2023best} & 97.7 \\
			SignBERT+~(+R)~\cite{hu2023signbert+} & 97.8 \\
			\textit{Ours}~(+R) & \textbf{97.8} \\ \bottomrule
		\end{tabular}
	}
	\label{slr500}
\end{table}

\subsection{Different Evaluation Protocols}
In this section, we introduce more evaluation protocols to show the performance of our proposed method and provide a strong baseline for these protocols.

\textbf{Linear evaluation protocol.} For the linear evaluation protocol, we freeze the pre-trained encoder and add a learnable linear classifier layer after it. We utilize the whole training data to train the classifier with SGD optimizer~\cite{robbins1951stochastic}. We set the total training epochs to 70 and the initial learning rate to 0.01, then reduce it to 0.001 at epoch 50. The performance on this protocol is shown in Fig.~\ref{fig:linear}. Even if only the classifier is trained, our method still achieves promising performance in all datasets. Notably, in SLR500~\cite{huang2018attention} and NMFs\_CSL~\cite{hu2021global}, the recognition accuracy attains comparable results to that of the fine-tuned whole framework, which demonstrates the discriminative capacity of the learned representations after pre-training.

\begin{figure}[t]
	\centering
	\includegraphics[width=1.0\linewidth]{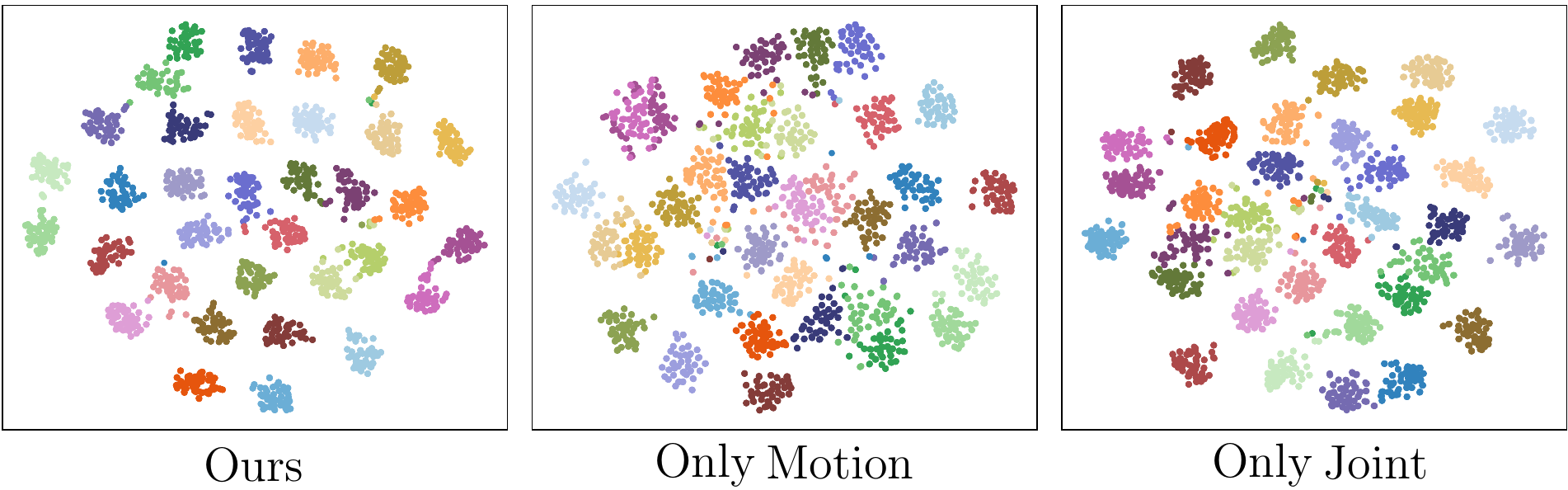}
	\caption{t-SNE~\cite{van2008visualizing} visualization of feature embeddings. We sample 34 sign words from SLR500 dataset and visualize the features extracted from our proposed method and single branches, denoted as ``Ours", ``Only Motion" and ``Only Joint", respectively.
	}
	\label{fig:tsne}
\end{figure}

\textbf{Semi-supervised evaluation protocol.} In the semi-supervised evaluation protocol, both labeled and unlabeled data are included during training. It aims to train the framework with only partially labeled data to achieve comparable performance with the one trained with the whole labeled data. As shown in Fig.~\ref{fig:semi}, we randomly select a proportion of labeled data to fine-tune the pre-trained framework. It is observed that as the proportion of labeled data increases, the performance of each dataset grows at a progressively slower pace.
Finally, we found that utilizing about 60\% labeled data could achieve comparable performance with the whole labeled data. This result also demonstrates the remarkable effectiveness of our pre-training framework.

\textbf{Comparison with other self-supervised methods.} We compare other self-supervised pre-training methods in the SLR domain, \textit{i.e.,} SignBERT~\cite{hu2021signbert}, SignBERT+~\cite{hu2023signbert+} and BEST~\cite{zhao2023best} under both evaluation protocols, respectively. As shown in Tab.~\ref{fig:linear} and Tab.~\ref{fig:semi}, our proposed method shows significant performance gains across various evaluation protocols, which is attributed to learning consistent and discriminative representation during the pre-training stage. In contrast, previous pre-training methods generally learn frame-wise features and overlook effectively holistic information present in sign pose sequences, leading to inferior performance. In comparison to these methods, our proposed method demonstrates the effectiveness of learning discriminative representation with designed spatial-temporal consistency during pre-training. 

\begin{table*}[!t]
	\centering
	\small
	\tabcolsep=4pt
	\caption{The performance of different pre-training methods under the semi-supervised evaluation protocol on four benchmarks and their subsets. ``semi-supervised" denote the evaluation protocol. The ``percent" denotes the proportion of labeled training data utilized during the fine-tuning stage.}
	\begin{center}
		\resizebox{\linewidth}{!}{
			\begin{tabular}{c|c|c|cc|cc|cc|cc|cc|cc|c|c}
				\toprule
				\multirow{2}{*}[-0.05in]{\makecell[c]{Protocol}} & \multirow{2}{*}[-0.05in]{\makecell[c]{\makecell[c]{Percent}}} &
				\multirow{2}{*}[-0.05in]{\makecell[c]{\makecell[c]{Methods}}} & \multicolumn{2}{c|}{MSASL100} & \multicolumn{2}{c|}{MSASL200} & \multicolumn{2}{c|}{MSASL1000} & \multicolumn{2}{c|}{WLASL100} & \multicolumn{2}{c|}{WLASL300} & \multicolumn{2}{c|}{WLASL2000} & \multirow{2}{*}[-0.05in]{\makecell[c]{NMFs-CSL}} & \multirow{2}{*}[-0.05in]{\makecell[c]{SLR500}}\\
				\cmidrule(lr){4-5} \cmidrule(lr){6-7} \cmidrule(lr){8-9} \cmidrule(lr){10-11} \cmidrule(lr){12-13} \cmidrule(lr){14-15}
				&   &  & P-I & P-C &  P-I &  P-C &  P-I &  P-C &  P-I &  P-C &  P-I &  P-C &  P-I &  P-C &  &\\ \midrule
				
				\multirow{4}{*}[-0.1in]{\makecell[c]{semi-\\supervised}} &\multirow{4}{*}[-0.1in]{ $20 \%$} & \multirow{1}{*}[-0.00in]{SignBERT~\cite{hu2021signbert}} & 52.31 & 52.42 & 38.78 & 38.92 & 19.28 & 19.36 & 24.88 & 25.02 & 22.26 & 22.85 & 10.37 & 10.41 & 43.3 & 57.6 \\
				\cmidrule(lr){3-3}
				& & \multirow{1}{*}[-0.00in]{SignBERT+~\cite{hu2023signbert+}} & 57.07 & 57.28 & 44.59& 44.75 & 23.44 & 24.15 & 31.40 & 32.28 & 26.95 & 27.60 & 14.11 & 13.27 & -- & 60.4 \\
				\cmidrule(lr){3-3}
				& & \multirow{1}{*}[-0.00in]{BEST~\cite{zhao2023best}} & 54.31 & 54.48 & 41.89 & 42.33 & 21.08 & 21.24 & 27.68 & 28.11 & 24.82 & 24.96 & 12.25 & 12.46 & 45.0 & 59.5  \\
				\cmidrule(lr){3-3}
				& & \multirow{1}{*}[-0.00in]{Ours} & \textbf{71.07} & \textbf{72.06} & \textbf{65.12} & \textbf{65.04} & \textbf{39.09} & \textbf{35.07} & \textbf{49.61} & \textbf{49.87} & \textbf{41.02} & \textbf{41.22} & \textbf{21.02} & \textbf{17.89} & \textbf{66.3} & \textbf{95.2} \\
				\midrule
				\multirow{4}{*}[-0.1in]{\makecell[c]{semi-\\supervised}} &\multirow{4}{*}[-0.1in]{ $40 \%$} & \multirow{1}{*}[-0.00in]{SignBERT~\cite{hu2021signbert}} & 62.47 & 63.24 & 60.08 & 61.13 & 35.36 & 32.12 & 55.77 & 56.03 & 42.53 & 43.66 & 23.54 & 21.47 & 52.7 & 82.1 \\
				\cmidrule(lr){3-3}
				& & \multirow{1}{*}[-0.00in]{SignBERT+~\cite{hu2023signbert+}} & 68.65 & 69.84 & 64.81 & 65.26 & 45.96 & 43.78 & 60.30 & 60.44 & 49.35 & 49.79 & 29.82 & 27.49 & - & 83.4 \\
				\cmidrule(lr){3-3}
				& & \multirow{1}{*}[-0.00in]{BEST~\cite{zhao2023best}} & 64.43 & 65.71 & 62.31 & 62.98 & 42.12 & 40.41 & 57.69 & 58.32 & 44.31 & 45.53 & 26.18 & 24.88 & 53.5 & 82.6 \\
				\cmidrule(lr){3-3}
				& & \multirow{1}{*}[-0.00in]{Ours} & \textbf{80.85} & \textbf{80.83} & \textbf{74.54} & \textbf{74.94}  & \textbf{51.49} & \textbf{47.41} & \textbf{72.48} & \textbf{73.97} & \textbf{62.57} & \textbf{63.09} & \textbf{36.52} & \textbf{32.49} & \textbf{71.2} & \textbf{95.9} \\
				\midrule
				\multirow{4}{*}[-0.1in]{\makecell[c]{semi-\\supervised}} &\multirow{4}{*}[-0.1in]{ $60 \%$} & \multirow{1}{*}[-0.00in]{SignBERT~\cite{hu2021signbert}} & 69.48 & 70.03 & 67.13 & 67.72 & 41.28 & 38.45 & 64.68 & 65.43 & 52.17 & 52.74 & 31.84 & 28.53 & 59.8 & 88.9 \\
				\cmidrule(lr){3-3}
				& & \multirow{1}{*}[-0.00in]{SignBERT+~\cite{hu2023signbert+}} & 75.52 & 75.97 & 71.63 & 72.16 & 50.35 & 47.82 & 68.89 & 69.74 & 61.93 & 62.41 & 37.58 & 34.64 & - & 89.7\\
				\cmidrule(lr){3-3}
				& & \multirow{1}{*}[-0.00in]{BEST~\cite{zhao2023best}} & 72.28 & 72.86 & 68.85 & 69.24& 48.27& 46.58 & 66.24 & 66.97 & 55.79 & 56.36 & 35.26 & 33.58 & 60.4 & 89.4 \\
				\cmidrule(lr){3-3}
				& & \multirow{1}{*}[-0.00in]{Ours} & \textbf{82.56} & \textbf{82.88} & \textbf{79.76} & \textbf{80.57}  & \textbf{59.06} & \textbf{56.42} & \textbf{82.17} & \textbf{82.80} & \textbf{70.66} & \textbf{71.16} & \textbf{44.37} & \textbf{42.76} & \textbf{73.6} & \textbf{96.3} \\
				\bottomrule     
			\end{tabular}
        }
	\end{center}
	\label{fig:semi}
\end{table*}

\subsection{Ablation Study}
In this section, we conduct ablation study to validate the effectiveness of our approach and select proper hyper-parameters. 
For fair comparison, experiments are mainly performed on the MSASL dataset and we report the top-1 accuracy under the per-instance and per-class metrics as the indicator.

\textbf{Pre-training data scale.} In Tab.~\ref{data_scale}, we investigate the effect of the pre-training data scale. The first row denotes the proposed framework is performed without pre-training.  It is observed that the performance gradually increases with the increment in the proportion of the pre-training data scale. The result demonstrates that our method is promising to the pre-training for large-scale data.

\begin{table}[t!]
	\small
	\tabcolsep=9pt
	\caption{Impact of the data scale during pre-training on MSASL dataset. The ``Percent" denotes the proportion of \\ pre-training data.}
	\begin{center}
		\resizebox{\linewidth}{!}{
			\begin{tabular}{c|cc|cc|cc}
				\toprule
				\multicolumn{1}{c|}{\multirow{2}{*}{Percent}} & \multicolumn{2}{c|}{MSASL100} & \multicolumn{2}{c|}{MSASL200} & \multicolumn{2}{c}{MSASL1000} \\
				\cmidrule(lr){2-3} \cmidrule(lr){4-5} \cmidrule(lr){6-7}
				& P-I & P-C & P-I & P-C & P-I & P-C  \\ \midrule
				0\% & 69.88  & 70.97 & 66.89 & 67.75& 51.37 & 47.63  \\
				25\% &80.32 & 80.99 & 75.28 & 76.33 & 59.06 & 57.04 \\
				50\% & 83.22 & 83.75 & 79.62 & 80.68 & 62.82 & 60.31 \\
				75\% & 84.64 & 85.19 & 80.24 & 81.08 & 64.57 & 62.24 \\
				100\% &\textbf{86.26} & \textbf{86.63}  &\textbf{82.93} & \textbf{83.43}   & \textbf{65.22} & \textbf{62.68}    \\ \bottomrule  
			\end{tabular}
		}
	\end{center}
	\label{data_scale}
\end{table}

\begin{table}[!t]
	\centering
	\footnotesize
	\tabcolsep=5pt
	\caption{Impact of the knowledge transfer module on MSASL dataset. ``w $KT$" denotes utilizing knowledge transfer during pre-training, while ``w/o $KT$" denotes pre-training without the proposed knowledge transfer module.}
	\begin{center}
		\resizebox{\linewidth}{!}{
			\begin{tabular}{c|c|cc|cc|cc}
				\toprule
				\multirow{2}{*}[-0.05in]{\makecell[c]{Modality}} & \multirow{2}{*}[-0.05in]{\makecell[c]{\makecell[c]{Knowledge \\ Transfer}}} & \multicolumn{2}{c|}{MSASL100} & \multicolumn{2}{c|}{MSASL200} & \multicolumn{2}{c}{MSASL1000} \\
				\cmidrule(lr){3-4} \cmidrule(lr){5-6} \cmidrule(lr){7-8}
				&   & P-I & P-C &  P-I &  P-C &  P-I &  P-C\\ \midrule  
				\multirow{2}{*}[-0.0in]{\makecell[c]{Joint}} &\multirow{1}{*}[-0.00in]{ w/o $KT$}& 81.79 & 82.13  & 76.64 & 77.17 & 59.22 & 57.41  \\
				\cline{2-2}
				& \multirow{1}{*}[-0.02in]{ w $KT$} & \textbf{83.75} & \textbf{83.88}  & \textbf{80.57} & \textbf{80.93} & \textbf{61.07} & \textbf{59.39} \\ \midrule
				\multirow{2}{*}[-0.00in]{\makecell[c]{Motion}} & \multirow{1}{*}[-0.00in]{ w/o $KT$} & 79.74 & 79.91  & 73.83 & 74.26 & 56.63 & 53.95 \\
				\cline{2-2}
				& \multirow{1}{*}[-0.02in]{ w $KT$} & \textbf{81.51} & \textbf{81.60}  & \textbf{75.86} & \textbf{76.00} & \textbf{58.58} & \textbf{55.85} \\
				\bottomrule     
			\end{tabular}
		}
	\end{center}
	\label{Fine-tune}
\end{table}

\textbf{Pre-training with different modalities.} In the first row of Tab.~\ref{Pre-Train}, we pre-train our framework with three settings: only joint branch, only motion branch and jointly both branches~(Ours). We set the performance under the setting of ``no pre-training" as the baseline. For the former two settings, the knowledge transfer module is not utilized. We also keep the same settings in the fine-tuning stage. Compared with the single modality, unifying both modalities boosts the performance in different scale datasets. Moreover, in order to display the comprehensive representation of our proposed method, we directly visualize the feature embeddings learned from pre-training under three different settings with t-SNE~\cite{van2008visualizing}. The final results are illustrated in Fig.~\ref{fig:tsne}. For both joint and motion modalities, the representation of our proposed method is more compactly clustered than each of them. The result also validates the powerful capability of our method.

\textbf{Pre-training with different pose style.} We further conduct experiment on the different pose styles, \textit{i.e.,} ``separate Hand \& Trunk" and ``unified Hand \& Trunk". The former denotes extracting the features of hand and trunk poses separately, while the latter denotes extracting them as a single feature. As shown in Tab.~\ref{Pre-Train}, the unified feature of hand and trunk poses causes the degradation of performance in different modal settings. We argue that the decoupled hand and trunk features could explicitly mine more sufficient cues from different granularity and further enhance the holistic representation of sign pose sequence via the consistency constraint. These results validate that the fine-grained hand and coarse-grained trunk poses can effectively improve performance.

	\begin{table*}[!t]
		\centering
		\tabcolsep=20pt
		\caption{Impact of different modalities and different pose styles during pre-training on MSASL dataset. ``Only Joint" and ``Only Motion" denote that our method is pre-trained with joint and motion modality, respectively. ``Joint+Motion" indicates pre-training our method with both modalities. ``Unified Hand \& Trunk" denotes to extract them as a single feature, while ``Separate Hand \& Trunk" denotes to extract the features of hand and trunk poses separately.}
		\begin{center}
			\resizebox{\linewidth}{!}{
					\begin{tabular}{c|c|cc|cc|cc}
						\toprule
						\multirow{2}{*}[-0.05in]{\makecell[c]{Pose Style}}  & \multirow{2}{*}[-0.05in]{\makecell[c]{Pre-Train \\ Modality}} & \multicolumn{2}{c|}{MSASL100} & \multicolumn{2}{c|}{MSASL200} & \multicolumn{2}{c}{MSASL1000} \\
						\cmidrule(lr){3-4} \cmidrule(lr){5-6} \cmidrule(lr){7-8}
						& & P-I & P-C &  P-I &  P-C &  P-I &  P-C\\ \midrule 
						\multirow{3}{*}[-0.0in]{\makecell[c]{Unified \\ Hand \& Trunk}} & Only Joint  & 81.32  & 81.54 &  77.35  & 77.89  & 58.54 & 55.74 \\
						& Only Motion  & 79.35 & 79.67  & 71.75  & 72.26  & 55.41 & 52.65 \\
						& Joint+Motion & \textbf{82.95} & \textbf{83.31} & \textbf{79.29} & \textbf{80.16} & \textbf{60.37} & \textbf{58.23} \\ 
						\midrule
						\multirow{4}{*}[-0.0in]{\makecell[c]{Separate \\ Hand \& Trunk}} 
						& w/o Pretrain & 72.31 & 72.43 & 66.81 & 67.48 & 52.26 & 50.10 \\
						& Only Joint  & 83.75 & 83.88  & 80.57 & 80.93 & 61.07 & 59.39  \\
						& Only Motion  & 81.51 & 81.60  & 75.86 & 76.00 & 58.58 & 55.85 \\
						& Joint+Motion &\textbf{86.26} & \textbf{86.63}  &\textbf{82.93} & \textbf{83.43}   & \textbf{65.22} & \textbf{62.68}  \\
						\bottomrule
						
					\end{tabular}
				}
		\end{center}
		\label{Pre-Train}
	\end{table*}

\begin{table}[t]
	\centering
	\tabcolsep=7pt
	\caption{Impact of the different granularity information on MSASL datasets. ``Hand" and ``Trunk" denote the hand gestures and trunk pose. The last row denotes our proposed method.}
	\begin{center}
		\resizebox{\linewidth}{!}{
			\begin{tabular}{cc|cccccc}
				\toprule
				\multicolumn{2}{c|}{Granularity} & \multicolumn{2}{c}{MSASL100} & \multicolumn{2}{c}{MSASL200} & \multicolumn{2}{c}{MSASL1000} \\
				\cmidrule(lr){3-4} \cmidrule(lr){5-6} \cmidrule(lr){7-8}
				Hand   &  Trunk    & P-I & P-C &  P-I &  P-C &  P-I &  P-C\\ \midrule
				\checkmark &  & 82.96 & 83.36 & 79.10 & 79.27 & 58.05 & 55.37 \\
				& \checkmark & 55.61 & 53.82 & 44.81 & 44.96 & 26.97 & 24.12 \\
				\checkmark & \checkmark & \textbf{86.26} & \textbf{86.63} &\textbf{82.93}  & \textbf{83.43}  & \textbf{65.22}  & \textbf{62.68} \\ \bottomrule      
			\end{tabular}
		}
	\end{center}
	\label{design}
\end{table}


\begin{table}[!h]
	\centering
	\tabcolsep=8pt
	\caption{Impact of the proposed objectives on MSASL dataset. $\mathcal{L}_{con}$ includes $\mathcal{L}_{con}^{J}$ and $\mathcal{L}_{con}^M$ in two branches. The first row denotes a baseline only with contrastive loss.}
	\begin{center}
		\resizebox{\linewidth}{!}{
			\begin{tabular}{cc|cccccc}
				\toprule
				\multicolumn{2}{c|}{Objectives} & \multicolumn{2}{c}{MSASL100} & \multicolumn{2}{c}{MSASL200} & \multicolumn{2}{c}{MSASL1000} \\
				\cmidrule(lr){3-4} \cmidrule(lr){5-6} \cmidrule(lr){7-8}
				$\mathcal{L}_{con}$      &  $\mathcal{L}_{KT}$     & P-I & P-C &  P-I &  P-C &  P-I &  P-C\\ \midrule
				&            & 78.29 & 78.65 & 75.35 & 75.72 & 57.32 & 54.25 \\
				\checkmark &  & 81.01 & 81.45 & 78.51 & 78.92 & 61.74 & 59.14  \\
				& \checkmark & 84.94 & 85.32 & 81.46 & 81.93 & 64.07 & 62.18 \\
				\checkmark & \checkmark & \textbf{86.26} & \textbf{86.63}  &\textbf{82.93} & \textbf{83.43}   & \textbf{65.22} & \textbf{62.68} \\ 
				\bottomrule        
			\end{tabular}
		}
	\end{center}
	\label{objective}
\end{table}

\textbf{Impact of knowledge transfer for single modality.} In Tab.~\ref{Fine-tune}, we conduct experiments to validate the effectiveness of the knowledge transfer module. The ``w $KT$'' means that we first pre-train our method with both modalities, then fine-tune it with a single modality. 
By combining knowledge from both joint and motion modalities, the representation of each individual modality is enhanced, resulting in better performance compared to relying solely on information from a single modality.
Concretely, the performance of only utilizing the joint modality improves 1.96\%, 3.93\% and 1.85\% accuracy under the Top-1 per-instance metric.
It also proves that both modalities contain complementary information to learn more semantic representation.

\begin{figure}[!t]
	\centering
	\includegraphics[width=0.9\linewidth]{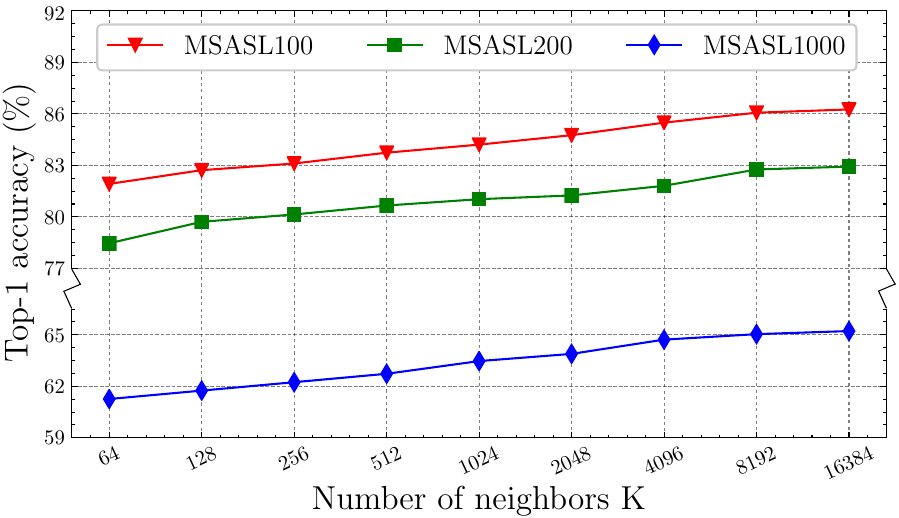}
	\caption{Impact of the number of neighbors $K$ in knowledge transfer module on MSASL dataset. The horizontal axis indicates the value of $K$, while the vertical axis indicates the top-1 accuracy.}
	\label{fig:neighbors}
\end{figure}

\begin{table*}[!h]
    \centering
    \tabcolsep=6pt
    \caption{Impact of the loss weights on different datasets. $\lambda_J$ and $\lambda_M$ denote the proportion of the two branch weights. The equal proportion of both weights is beneficial to our framework.}
    \begin{center}
        \resizebox{\linewidth}{!}{
            \begin{tabular}{cc|cccccc|cccccc|c|c}
                \toprule
                \multicolumn{2}{c|}{Weights} & \multicolumn{2}{c}{MSASL100} & \multicolumn{2}{c}{MSASL200} & \multicolumn{2}{c|}{MSASL1000} & \multicolumn{2}{c}{WLASL100} & \multicolumn{2}{c}{WLASL300} & \multicolumn{2}{c|}{WLASL2000}& \multicolumn{1}{c|}{NMFs-CSL} & \multicolumn{1}{c}{SLR500} \\
                \cmidrule(lr){3-4} \cmidrule(lr){5-6} \cmidrule(lr){7-8} \cmidrule(lr){9-10} \cmidrule(lr){11-12} \cmidrule(lr){13-14} \cmidrule(lr){15-15} \cmidrule(lr){16-16}   
                $\lambda_J$      & 	$\lambda_M$      & P-I & P-C &  P-I &  P-C &  P-I &  P-C & P-I & P-C &  P-I &  P-C &  P-I &  P-C & P-I & P-I \\ \midrule
                0.3 &  0.7 & 81.77& 82.26& 77.28& 77.87 & 61.10 & 58.61 & 82.34 & 82.86 & 72.86 & 73.17 & 47.79 & 46.28 & 72.4 & 94.5 \\
                0.4 &  0.6 & 82.71 & 82.96 & 78.84 & 79.29 & 62.09 & 59.51 & 83.28 & 83.71 & 74.53 & 75.17 &49.22 & 47.68 & 73.3 & 95.8 \\
                0.5 & 0.5 & \textbf{86.26} & \textbf{86.63} &\textbf{82.93}  & \textbf{83.43}  & \textbf{65.22}  & \textbf{62.68}  & \textbf{86.43} & \textbf{87.13} & \textbf{77.84} & \textbf{78.42} & \textbf{51.98} & \textbf{49.46} & \textbf{75.6} & \textbf{96.9}\\
                0.6 & 0.4 & 82.87 & 83.26 & 79.23 & 79.76 & 62.17 & 59.61 & 83.84 & 84.15 & 75.06 & 75.79 &  49.08 & 47.44 & 73.8 & 95.6 \\ 
                0.7 & 0.3 & 81.41& 81.75 & 76.85 & 77.22 & 60.94& 58.29 & 82.53 & 82.94 & 73.24 & 73.82 & 48.31 & 46.57 & 72.7 & 94.7 \\ 
                \bottomrule        
            \end{tabular}
        }
    \end{center}
    \label{weights}
\end{table*}

\begin{figure*}[!t]
	\centering
	\includegraphics[width=0.9\linewidth]{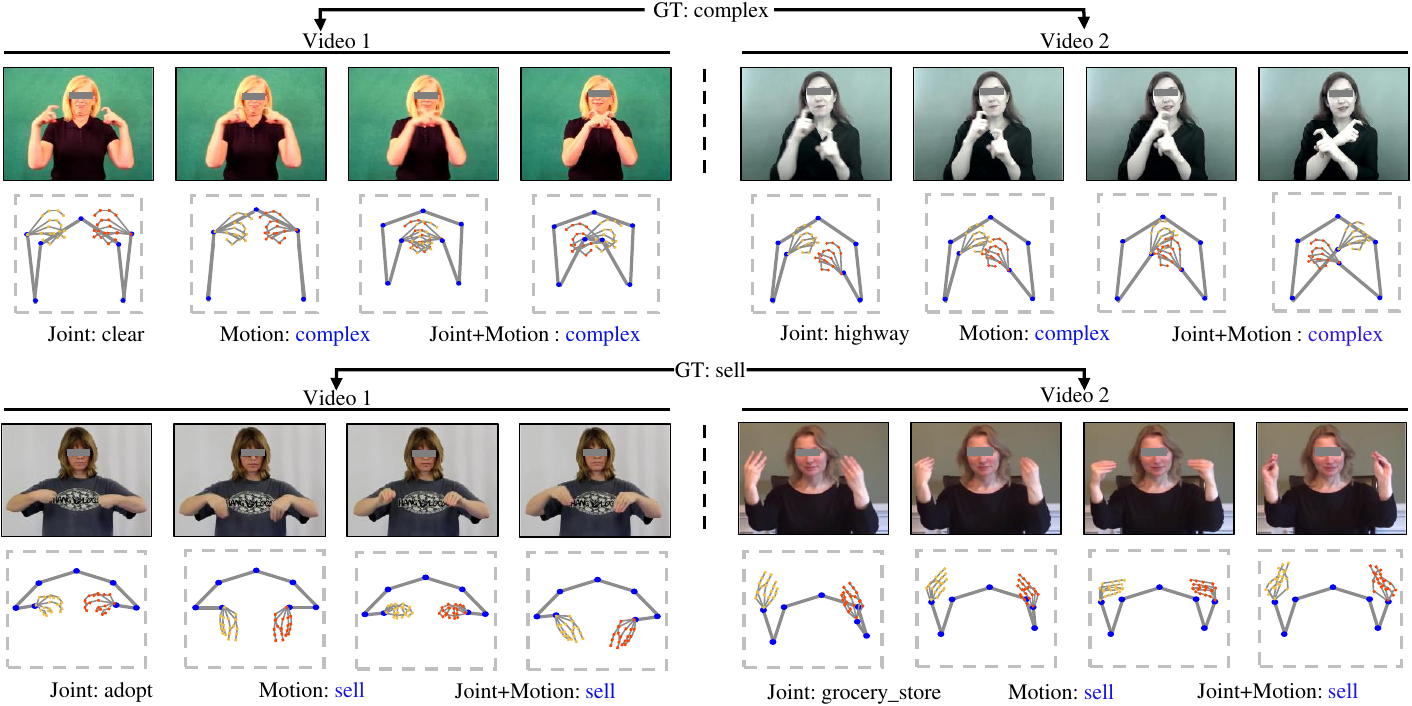}
	\caption{Qualitative results of the effectiveness of different order information. We visualize several samples that the different signers perform the same sign words. Due to the variability among human characteristics, the static joint modality of sign pose sequences occasionally causes the wrong prediction results. Different from it, the dynamic motion among different signers is consistent and involves the complementary representation to the joint modality. The combination of both modalities predicts better results.}
	\label{fig:motion}
\end{figure*}

\begin{figure*}[!t]
	\centering
	\includegraphics[width=0.9\linewidth]{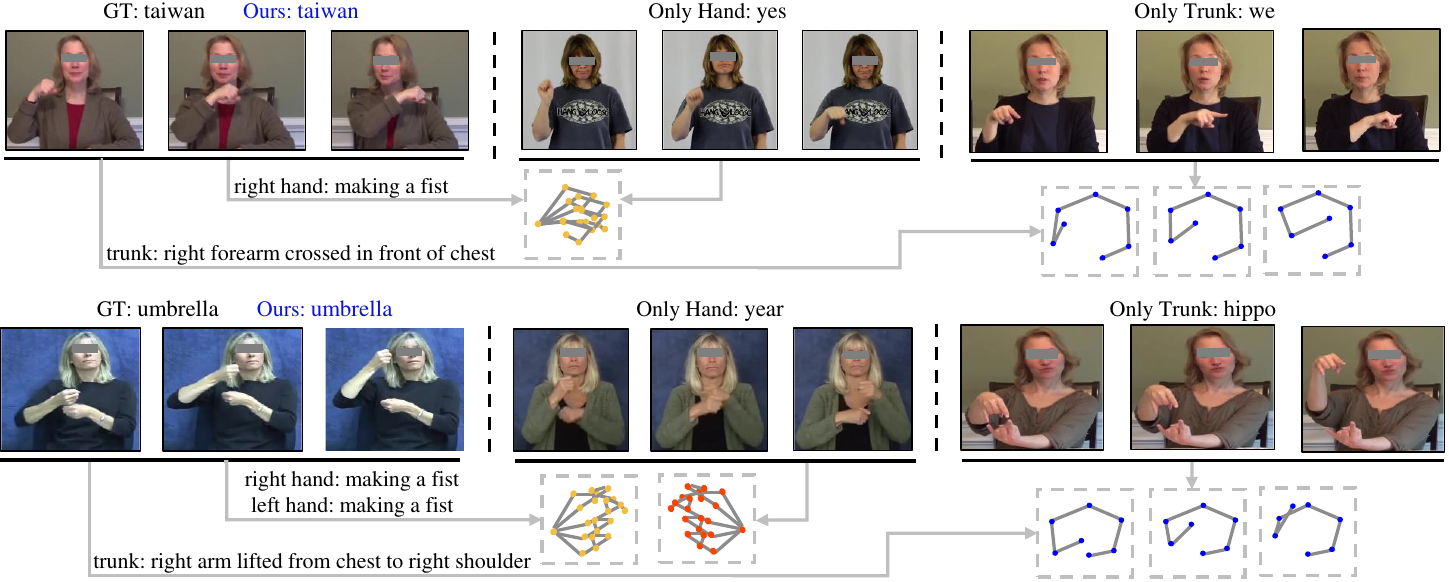}
	\caption{Qualitative results of the effectiveness of the collaboration between hand and trunk. We visualize the key frames in the corresponding video of the predicted word under three settings, \textit{i.e.,} ``Ours", ``Only Hand" and ``Only Trunk". It is noted that the close collaboration of hand and trunk is essential to correct the errors of utilizing hand or trunk pose alone. The red color denotes the correct classification.}
	\label{fig:hand_trunk}
\end{figure*}

\textbf{Number of neighbors $K$.} The number of neighbors controls the abundance of reliable information utilized in the knowledge transfer module. As shown in Fig.~\ref{fig:neighbors}, we explore the performance of  the learned representation with respect to the number of neighbors $K$. The performance gradually increases with the increment of $K$. When $K$ is large enough~($K \geq 8192$), the accuracy becomes saturated. The reason of this condition is that the newly added neighbors are far away from the center embedding and provide little effective knowledge. This result validates that reliable knowledge transfer is essential to representation learning.

\begin{table}[!t]
	\small
	\tabcolsep=5pt
	\caption{Impact of different fusion strategies on MSASL dataset. ``Avg." is the abbreviation of ``Average".}
	\begin{center}
		\resizebox{\linewidth}{!}{
				\begin{tabular}{c|cc|cc|cc}
					\toprule
					\multicolumn{1}{c|}{\multirow{2}{*}{Fusion Strategy}} & \multicolumn{2}{c|}{MSASL100} & \multicolumn{2}{c|}{MSASL200} & \multicolumn{2}{c}{MSASL1000} \\
					\cmidrule(lr){2-3} \cmidrule(lr){4-5} \cmidrule(lr){6-7}
					& P-I & P-C & P-I & P-C & P-I & P-C  \\ \midrule
					Late Fusion & 90.76  & 91.97 & 88.82 & 89.22 & 74.78 & 72.18  \\
					Early Fusion-Summation & 92.43  &  92.61 & 89.84  & 90.23  & 76.45  & 73.93 \\ 
					Early Fusion-Adaptive Avg. & 92.69 & 93.07 &  90.15 &  90.67 & 76.74  &  74.11 \\
					Early Fusion-Concatenation &  \textbf{93.74} & \textbf{93.95} & \textbf{91.35} & \textbf{91.79}  &  \textbf{77.35} & \textbf{74.19}  \\
					\bottomrule  
				\end{tabular}
			}
	\end{center}
	\label{fusion_strategy}
\end{table}

\textbf{Impact of different granularity information.} In Tab.~\ref{design}, we separately fine-tune our framework with hand and trunk features and report the performance in the first two rows. It is observed that the performance of only utilizing trunk data exists a large margin compared with using hand data. We consider the reason that the fine-grained hand gestures contain more useful information than the coarse-grained trunk pose. The prediction accuracy achieves better improvement via leveraging both granularity information.

\textbf{Impact of proposed objectives.} As shown in Tab.~\ref{objective}, we compare the impact of proposed objectives, \textit{i.e.,} intra-modal consistency constraint and cross-modal knowledge transfer constraint. The first row shows the performance of pre-training our method only with common contrastive loss. It is observed that each objective individually improves the performance. As leveraging both objectives, our method achieves remarkable performance gain, \textit{i.e.,} 7.97\%, 7.58\% and 7.90\% for per-instance accuracy improvements.

\textbf{Impact of the loss weights.} In Tab.~\ref{weights}, we further study the impact of the loss weights, \textit{i.e.,} $\lambda_J$ and $\lambda_M$. Concretely, we keep the summation of the two weights at 1.0 and adjust their different percentages. It is noted that equal utilization of joint and motion branches leads to optimal performance.

\textbf{Different fusion strategies.} We explore the diverse fusion strategies with our pre-trained skeleton-based model and another RGB-based model I3D~\cite{carreira2017quo}, as shown in Tab.~\ref{fusion_strategy}. Concretely, in addition to late fusion, we employ various merging strategies for the early fusion of features, \textit{i.e.,} summation, concatenation and adaptive average. It is observed that the early fusion strategy is overall better than late fusion strategy. This result is facilitated by the joint optimization of the RGB-based model and our pre-trained model during early fusion. Moreover, different fusion strategies show various performances on MSASL dataset. The concatenation operation achieves the best performance due to the independent preservation of features from the different modalities and avoiding information overlapping.

\subsection{Qualitative Visualization}
We provide more qualitative results to demonstrate the feasibility and effectiveness of our proposed method. As shown in Fig.~\ref{fig:hand_trunk}, we visualize the prediction results of our method, only adopting hand or trunk, respectively. The similar hand or trunk pose sequences in different sign words lead to incorrect results, \textit{i.e.,} right hand making a fist in \textit{taiwan} and \textit{yes} words. Our approach collaborates spatial information, \textit{i.e,} hand and trunk, to achieve accurate predictions. 
Moreover, we display the results of the same sign word performed by different signers in Fig.~\ref{fig:motion}. It is observed that the variances of joint information from different signers lead to wrong predictions. Different from the joint modality, the dynamic sign pose motion contains invariant temporal features among different signers to correct wrong prediction results, which demonstrates the indispensable of the motion modality.

\section{Conclusion}

In this paper, we propose a self-supervised pre-training framework to learn instance discriminative representation for sign language recognition in a contrastive learning paradigm. 
Specifically, we excavate sufficient information via spatial-temporal consistency from two perspectives.
On one hand, we integrate fine-grained hand gestures and coarse-grained trunk poses to represent holistic sign language meaning. On the other hand, considering the complementarity between joint and motion modalities, we explicitly utilize motion information to fertilize the comprehensive representation. Furthermore, we design a knowledge transfer module to convey reliable information between joint and motion modalities to enhance the representative capacity. Extensive experiments validate the effectiveness of our method, achieving new state-of-the-art performance.

\bibliographystyle{IEEEtran}
\bibliography{main.bib}

\begin{thebibliography}{10}
\providecommand{\url}[1]{#1}
\csname url@samestyle\endcsname
\providecommand{\newblock}{\relax}
\providecommand{\bibinfo}[2]{#2}
\providecommand{\BIBentrySTDinterwordspacing}{\spaceskip=0pt\relax}
\providecommand{\BIBentryALTinterwordstretchfactor}{4}
\providecommand{\BIBentryALTinterwordspacing}{\spaceskip=\fontdimen2\font plus
\BIBentryALTinterwordstretchfactor\fontdimen3\font minus \fontdimen4\font\relax}
\providecommand{\BIBforeignlanguage}[2]{{%
\expandafter\ifx\csname l@#1\endcsname\relax
\typeout{** WARNING: IEEEtran.bst: No hyphenation pattern has been}%
\typeout{** loaded for the language `#1'. Using the pattern for}%
\typeout{** the default language instead.}%
\else
\language=\csname l@#1\endcsname
\fi
#2}}
\providecommand{\BIBdecl}{\relax}
\BIBdecl

\bibitem{li2020word}
D.~Li, C.~Rodriguez, X.~Yu, and H.~Li, ``Word-level deep sign language recognition from video: {A} new large-scale dataset and methods comparison,'' in \emph{WACV}, 2020, pp. 1459--1469.

\bibitem{li2020transferring}
D.~Li, X.~Yu, C.~Xu, L.~Petersson, and H.~Li, ``Transferring cross-domain knowledge for video sign language recognition,'' in \emph{CVPR}, 2020, pp. 6205--6214.

\bibitem{koller2018deep}
O.~Koller, S.~Zargaran, H.~Ney, and R.~Bowden, ``Deep sign: Enabling robust statistical continuous sign language recognition via hybrid {CNN-HMMs},'' \emph{IJCV}, vol. 126, no.~12, pp. 1311--1325, 2018.

\bibitem{joze2018ms}
H.~R.~V. Joze and O.~Koller, ``{MS-ASL}: A large-scale data set and benchmark for understanding american sign language,'' in \emph{BMVC}, 2019, pp. 1--16.

\bibitem{tunga2021pose}
A.~Tunga, S.~V. Nuthalapati, and J.~Wachs, ``Pose-based sign language recognition using {GCN} and {BERT},'' in \emph{WACV}, 2021, pp. 31--40.

\bibitem{selvaraj-etal-2022-openhands}
P.~Selvaraj, G.~Nc, P.~Kumar, and M.~Khapra, ``{O}pen{H}ands: Making sign language recognition accessible with pose-based pretrained models across languages,'' in \emph{ACL}, 2022, pp. 2114--2133.

\bibitem{albanie2020bsl}
S.~Albanie, G.~Varol, L.~Momeni, T.~Afouras, J.~S. Chung, N.~Fox, and A.~Zisserman, ``{BSL-1K}: Scaling up co-articulated sign language recognition using mouthing cues,'' in \emph{ECCV}, 2020, pp. 35--53.

\bibitem{hu2021signbert}
H.~Hu, W.~Zhao, W.~Zhou, Y.~Wang, and H.~Li, ``{SignBERT}: Pre-training of hand-model-aware representation for sign language recognition,'' in \emph{ICCV}, 2021, pp. 11\,087--11\,096.

\bibitem{zhao2023best}
W.~Zhao, H.~Hu, W.~Zhou, J.~Shi, and H.~Li, ``{BEST}: Bert pre-training for sign language recognition with coupling tokenization,'' \emph{arXiv}, 2023.

\bibitem{devlin2018bert}
J.~Devlin, M.-W. Chang, K.~Lee, and K.~Toutanova, ``{BERT}: Pre-training of deep bidirectional transformers for language understanding,'' in \emph{NAACL}, 2019, pp. 4171--4186.

\bibitem{oord2018representation}
A.~v.~d. Oord, Y.~Li, and O.~Vinyals, ``Representation learning with contrastive predictive coding,'' \emph{arXiv}, 2018.

\bibitem{starner1995visual}
T.~E. Starner, ``Visual recognition of american sign language using hidden markov models.'' Massachusetts inst of tech Cambridge dept of brain and cognitive sciences, Tech. Rep., 1995.

\bibitem{ong2005automatic}
S.~C. Ong and S.~Ranganath, ``Automatic sign language analysis: A survey and the future beyond lexical meaning,'' \emph{IEEE TPAMI}, vol.~27, no.~06, pp. 873--891, 2005.

\bibitem{rastgoo2021sign}
R.~Rastgoo, K.~Kiani, and S.~Escalera, ``Sign language recognition: A deep survey,'' \emph{Expert Systems with Applications}, vol. 164, p. 113794, 2021.

\bibitem{hu2021hand}
H.~Hu, W.~Zhou, and H.~Li, ``Hand-model-aware sign language recognition,'' in \emph{AAAI}, vol.~35, 2021, pp. 1558--1566.

\bibitem{li2022transcribing}
D.~Li, C.~Xu, L.~Liu, Y.~Zhong, R.~Wang, L.~Petersson, and H.~Li, ``Transcribing natural languages for the deaf via neural editing programs,'' in \emph{AAAI}, 2022, pp. 11\,991--11\,999.

\bibitem{sincan2020autsl}
O.~M. Sincan and H.~Y. Keles, ``Autsl: A large scale multi-modal turkish sign language dataset and baseline methods,'' \emph{IEEE Access}, vol.~8, pp. 181\,340--181\,355, 2020.

\bibitem{bilge2022towards}
Y.~C. Bilge, R.~G. Cinbis, and N.~Ikizler-Cinbis, ``Towards zero-shot sign language recognition,'' \emph{IEEE TPAMI}, 2022.

\bibitem{koller2020quantitative}
O.~Koller, ``Quantitative survey of the state of the art in sign language recognition,'' \emph{arXiv}, 2020.

\bibitem{hu2023self}
L.~Hu, L.~Gao, Z.~Liu, and W.~Feng, ``Self-emphasizing network for continuous sign language recognition,'' in \emph{AAAI}, 2023, pp. 854--862.

\bibitem{niu2020stochastic}
Z.~Niu and B.~Mak, ``Stochastic fine-grained labeling of multi-state sign glosses for continuous sign language recognition,'' in \emph{ECCV}, 2020, pp. 172--186.

\bibitem{zuo2023natural}
R.~Zuo, F.~Wei, and B.~Mak, ``Natural language-assisted sign language recognition,'' in \emph{CVPR}, 2023, pp. 14\,890--14\,900.

\bibitem{xie2018rethinking}
S.~Xie, C.~Sun, J.~Huang, Z.~Tu, and K.~Murphy, ``Rethinking spatiotemporal feature learning: Speed-accuracy trade-offs in video classification,'' in \emph{ECCV}, 2018, pp. 305--321.

\bibitem{li2018co}
C.~Li, Q.~Zhong, D.~Xie, and S.~Pu, ``Co-occurrence feature learning from skeleton data for action recognition and detection with hierarchical aggregation,'' in \emph{IJCAI}, 2018, pp. 786--792.

\bibitem{yan2018spatial}
S.~Yan, Y.~Xiong, and D.~Lin, ``Spatial temporal graph convolutional networks for skeleton-based action recognition,'' in \emph{AAAI}, 2018.

\bibitem{9975251}
Z.~Li, X.~Gong, R.~Song, P.~Duan, J.~Liu, and W.~Zhang, ``{SMAM}: Self and mutual adaptive matching for skeleton-based few-shot action recognition,'' \emph{IEEE TIP}, vol.~32, pp. 392--402, 2023.

\bibitem{9997556}
Y.~Zhu, H.~Shuai, G.~Liu, and Q.~Liu, ``Multilevel spatial–temporal excited graph network for skeleton-based action recognition,'' \emph{IEEE TIP}, vol.~32, pp. 496--508, 2023.

\bibitem{9219176}
L.~Shi, Y.~Zhang, J.~Cheng, and H.~Lu, ``Skeleton-based action recognition with multi-stream adaptive graph convolutional networks,'' \emph{IEEE TIP}, vol.~29, pp. 9532--9545, 2020.

\bibitem{7450165}
Y.~Du, Y.~Fu, and L.~Wang, ``Representation learning of temporal dynamics for skeleton-based action recognition,'' \emph{IEEE TIP}, vol.~25, no.~7, pp. 3010--3022, 2016.

\bibitem{jiang2021skeletor}
T.~Jiang, N.~C. Camgoz, and R.~Bowden, ``Skeletor: Skeletal transformers for robust body-pose estimation,'' in \emph{CVPR}, 2021, pp. 3394--3402.

\bibitem{9523142}
S.~Jiang, B.~Sun, L.~Wang, Y.~Bai, K.~Li, and Y.~Fu, ``Skeleton aware multi-modal sign language recognition,'' in \emph{CVPRW}, 2021, pp. 3408--3418.

\bibitem{lee2023human}
T.~Lee, Y.~Oh, and K.~M. Lee, ``Human part-wise {3D} motion context learning for sign language recognition,'' in \emph{ICCV}, 2023, pp. 0--11.

\bibitem{kindiroglu2024transfer}
A.~A. Kindiroglu, O.~Kara, O.~Ozdemir, and L.~Akarun, ``Transfer learning for cross-dataset isolated sign language recognition in under-resourced datasets,'' \emph{arxiv}, 2024.

\bibitem{laines2023isolated}
D.~Laines, G.~Bejarano, M.~Gonzalez-Mendoza, and G.~Ochoa-Ruiz, ``Isolated sign language recognition based on tree structure skeleton images,'' in \emph{CVPRW}, 2023.

\bibitem{hu2023signbert+}
H.~Hu, W.~Zhao, W.~Zhou, and H.~Li, ``Signbert+: Hand-model-aware self-supervised pre-training for sign language understanding,'' \emph{IEEE TPAMI}, no.~01, pp. 1--20, 2023.

\bibitem{noroozi2016unsupervised}
M.~Noroozi and P.~Favaro, ``Unsupervised learning of visual representations by solving jigsaw puzzles,'' in \emph{ECCV}, 2016, pp. 69--84.

\bibitem{noroozi2018boosting}
M.~Noroozi, A.~Vinjimoor, P.~Favaro, and H.~Pirsiavash, ``Boosting self-supervised learning via knowledge transfer,'' in \emph{CVPR}, 2018, pp. 9359--9367.

\bibitem{gidaris2018unsupervised}
S.~Gidaris, P.~Singh, and N.~Komodakis, ``Unsupervised representation learning by predicting image rotations,'' in \emph{ICLR}, 2018, pp. 1--16.

\bibitem{He_2022_CVPR}
K.~He, X.~Chen, S.~Xie, Y.~Li, P.~Doll\'ar, and R.~Girshick, ``Masked autoencoders are scalable vision learners,'' in \emph{CVPR}, 2022, pp. 16\,000--16\,009.

\bibitem{tong2022videomae}
Z.~Tong, Y.~Song, J.~Wang, and L.~Wang, ``{VideoMAE}: Masked autoencoders are data-efficient learners for self-supervised video pre-training,'' \emph{arXiv}, 2022.

\bibitem{10246363}
Y.~Yang, B.~Wang, D.~Zhang, Y.~Yuan, Q.~Yan, S.~Zhao, Z.~You, and J.~Han, ``Self-supervised interactive embedding for one-shot organ segmentation,'' \emph{IEEE Transactions on Biomedical Engineering}, vol.~70, no.~10, pp. 2799--2808, 2023.

\bibitem{chen2020simple}
T.~Chen, S.~Kornblith, M.~Norouzi, and G.~Hinton, ``A simple framework for contrastive learning of visual representations,'' in \emph{ICML}, 2020, pp. 1597--1607.

\bibitem{zhao2021soda}
T.~Zhao, J.~Han, L.~Yang, B.~Wang, and D.~Zhang, ``{SODA}: Weakly supervised temporal action localization based on astute background response and self-distillation learning,'' \emph{International Journal of Computer Vision}, vol. 129, no.~8, pp. 2474--2498, 2021.

\bibitem{he2020momentum}
K.~He, H.~Fan, Y.~Wu, S.~Xie, and R.~Girshick, ``Momentum contrast for unsupervised visual representation learning,'' in \emph{CVPR}, 2020, pp. 9729--9738.

\bibitem{qian2021spatiotemporal}
R.~Qian, T.~Meng, B.~Gong, M.-H. Yang, H.~Wang, S.~Belongie, and Y.~Cui, ``Spatiotemporal contrastive video representation learning,'' in \emph{CVPR}, 2021, pp. 6964--6974.

\bibitem{pan2022learning}
J.~Pan, P.~Zhu, K.~Zhang, B.~Cao, Y.~Wang, D.~Zhang, J.~Han, and Q.~Hu, ``Learning self-supervised low-rank network for single-stage weakly and semi-supervised semantic segmentation,'' \emph{International Journal of Computer Vision}, vol. 130, no.~5, pp. 1181--1195, 2022.

\bibitem{sun2023masked}
X.~Sun, P.~Chen, L.~Chen, C.~Li, T.~H. Li, M.~Tan, and C.~Gan, ``Masked motion encoding for self-supervised video representation learning,'' in \emph{CVPR}, 2023, pp. 2235--2245.

\bibitem{chen2021rspnet}
P.~Chen, D.~Huang, D.~He, X.~Long, R.~Zeng, S.~Wen, M.~Tan, and C.~Gan, ``{RSPNet}: Relative speed perception for unsupervised video representation learning,'' in \emph{AAAI}, 2021, pp. 1045--1053.

\bibitem{ding2023prune}
S.~Ding, P.~Zhao, X.~Zhang, R.~Qian, H.~Xiong, and Q.~Tian, ``Prune spatio-temporal tokens by semantic-aware temporal accumulation,'' in \emph{CVPR}, 2023, pp. 16\,945--16\,956.

\bibitem{zeng2021graph}
R.~Zeng, W.~Huang, M.~Tan, Y.~Rong, P.~Zhao, J.~Huang, and C.~Gan, ``Graph convolutional module for temporal action localization in videos,'' \emph{IEEE TPAMI}, vol.~44, no.~10, pp. 6209--6223, 2021.

\bibitem{zeng2019graph}
------, ``Graph convolutional networks for temporal action localization,'' in \emph{ICCV}, 2019, pp. 7094--7103.

\bibitem{huang2021self}
L.~Huang, Y.~Liu, B.~Wang, P.~Pan, Y.~Xu, and R.~Jin, ``Self-supervised video representation learning by context and motion decoupling,'' in \emph{CVPR}, 2021, pp. 13\,886--13\,895.

\bibitem{feichtenhofer2021large}
C.~Feichtenhofer, H.~Fan, B.~Xiong, R.~Girshick, and K.~He, ``A large-scale study on unsupervised spatiotemporal representation learning,'' in \emph{CVPR}, 2021, pp. 3299--3309.

\bibitem{li20213d}
L.~Li, M.~Wang, B.~Ni, H.~Wang, J.~Yang, and W.~Zhang, ``{3D} human action representation learning via cross-view consistency pursuit,'' in \emph{CVPR}, 2021, pp. 4741--4750.

\bibitem{park2022probabilistic}
J.~Park, J.~Lee, I.-J. Kim, and K.~Sohn, ``Probabilistic representations for video contrastive learning,'' in \emph{CVPR}, 2022, pp. 14\,711--14\,721.

\bibitem{pan2021videomoco}
T.~Pan, Y.~Song, T.~Yang, W.~Jiang, and W.~Liu, ``Videomoco: Contrastive video representation learning with temporally adversarial examples,'' in \emph{CVPR}, 2021, pp. 11\,205--11\,214.

\bibitem{10100655}
J.~Wu, W.~Sun, T.~Gan, N.~Ding, F.~Jiang, J.~Shen, and L.~Nie, ``Neighbor-guided consistent and contrastive learning for semi-supervised action recognition,'' \emph{IEEE TIP}, vol.~32, pp. 2215--2227, 2023.

\bibitem{10109672}
W.~Lin, X.~Ding, Y.~Huang, and H.~Zeng, ``Self-supervised video-based action recognition with disturbances,'' \emph{IEEE TIP}, vol.~32, pp. 2493--2507, 2023.

\bibitem{huang2021ascnet}
D.~Huang, W.~Wu, W.~Hu, X.~Liu, D.~He, Z.~Wu, X.~Wu, M.~Tan, and E.~Ding, ``{ASCNet}: Self-supervised video representation learning with appearance-speed consistency,'' in \emph{ICCV}, 2021, pp. 8096--8105.

\bibitem{qing2022learning}
Z.~Qing, S.~Zhang, Z.~Huang, Y.~Xu, X.~Wang, M.~Tang, C.~Gao, R.~Jin, and N.~Sang, ``Learning from untrimmed videos: Self-supervised video representation learning with hierarchical consistency,'' in \emph{CVPR}, 2022, pp. 13\,821--13\,831.

\bibitem{wang2023masked}
R.~Wang, D.~Chen, Z.~Wu, Y.~Chen, X.~Dai, M.~Liu, L.~Yuan, and Y.-G. Jiang, ``Masked video distillation: Rethinking masked feature modeling for self-supervised video representation learning,'' in \emph{CVPR}, 2023, pp. 6312--6322.

\bibitem{hinton2015distilling}
G.~Hinton, O.~Vinyals, J.~Dean \emph{et~al.}, ``Distilling the knowledge in a neural network,'' \emph{arXiv}, vol.~2, no.~7, 2015.

\bibitem{park2019relational}
W.~Park, D.~Kim, Y.~Lu, and M.~Cho, ``Relational knowledge distillation,'' in \emph{CVPR}, 2019, pp. 3967--3976.

\bibitem{peng2019correlation}
B.~Peng, X.~Jin, J.~Liu, D.~Li, Y.~Wu, Y.~Liu, S.~Zhou, and Z.~Zhang, ``Correlation congruence for knowledge distillation,'' in \emph{ICCV}, 2019, pp. 5007--5016.

\bibitem{tung2019similarity}
F.~Tung and G.~Mori, ``Similarity-preserving knowledge distillation,'' in \emph{ICCV}, 2019, pp. 1365--1374.

\bibitem{mmpose2020}
M.~Contributors, ``{OpenMMLab} pose estimation toolbox and benchmark,'' \url{https://github.com/open-mmlab/mmpose}, 2020.

\bibitem{robbins1951stochastic}
H.~Robbins and S.~Monro, ``A stochastic approximation method,'' \emph{The annals of mathematical statistics}, pp. 400--407, 1951.

\bibitem{paszke2019pytorch}
A.~Paszke, S.~Gross, F.~Massa, A.~Lerer, J.~Bradbury, G.~Chanan, T.~Killeen, Z.~Lin, N.~Gimelshein, L.~Antiga \emph{et~al.}, ``{PyTorch}: An imperative style, high-performance deep learning library,'' in \emph{NeurIPS}, 2019, pp. 1--12.

\bibitem{zhao2024masa}
W.~Zhao, H.~Hu, W.~Zhou, Y.~Mao, M.~Wang, and H.~Li, ``{MASA}: Motion-aware masked autoencoder with semantic alignment for sign language recognition,'' \emph{IEEE TCSVT}, 2024.

\bibitem{carreira2017quo}
J.~Carreira and A.~Zisserman, ``Quo vadis, action recognition? a new model and the kinetics dataset,'' in \emph{CVPR}, 2017, pp. 6299--6308.

\bibitem{hu2021global}
H.~Hu, W.~Zhou, J.~Pu, and H.~Li, ``Global-local enhancement network for nmf-aware sign language recognition,'' \emph{ACM TOMM}, vol.~17, no.~3, pp. 1--19, 2021.

\bibitem{huang2018attention}
J.~Huang, W.~Zhou, H.~Li, and W.~Li, ``Attention-based {3D}-{CNNs} for large-vocabulary sign language recognition,'' \emph{IEEE TCSVT}, vol.~29, no.~9, pp. 2822--2832, 2018.

\bibitem{qiu2017learning}
Z.~Qiu, T.~Yao, and T.~Mei, ``Learning spatio-temporal representation with pseudo-{3D} residual networks,'' in \emph{ICCV}, 2017, pp. 5533--5541.

\bibitem{lin2019tsm}
J.~Lin, C.~Gan, and S.~Han, ``{TSM}: Temporal shift module for efficient video understanding,'' in \emph{ICCV}, 2019, pp. 7083--7093.

\bibitem{laptev2005space}
I.~Laptev, ``On space-time interest points,'' \emph{IJCV}, vol.~64, no.~2, pp. 107--123, 2005.

\bibitem{tang2015real}
A.~Tang, K.~Lu, Y.~Wang, J.~Huang, and H.~Li, ``A real-time hand posture recognition system using deep neural networks,'' \emph{ACM TIST}, vol.~6, no.~2, pp. 1--23, 2015.

\bibitem{van2008visualizing}
L.~Van~der Maaten and G.~Hinton, ``Visualizing data using t-sne.'' \emph{JMLR}, vol.~9, no.~11, 2008.

\end{thebibliography}

\vfill

\begin{IEEEbiography}[{\includegraphics[width=1in,height=1.25in,clip,keepaspectratio]{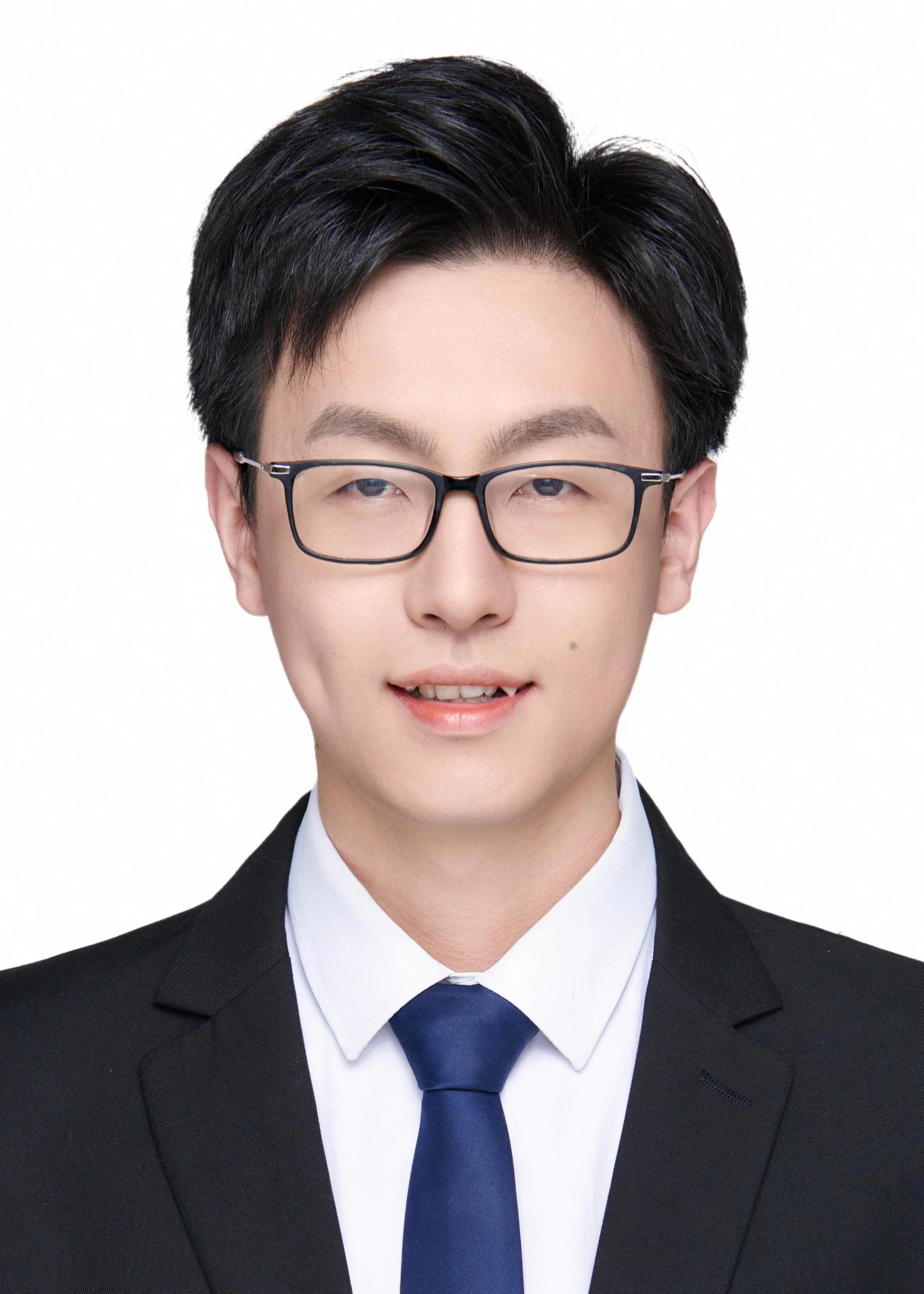}}]{Weichao Zhao} received B.E. degree in electronic information engineering from the University of Science and Technology of China~(USTC), and is currently pursuing the Ph.D. degree in data science with the School of Data Science from USTC.
His research interests include sign language understanding, self-supervised pre-training, multimodal representation learning and computer vision.
\end{IEEEbiography}

\begin{IEEEbiography}[{\includegraphics[width=1in,height=1.25in,clip,keepaspectratio]{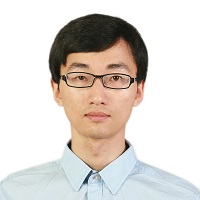}}]{Wengang Zhou} (S'20) received the B.E. degree in electronic information engineering from Wuhan University, China, in 2006, and the Ph.D. degree in electronic engineering and information science from the University of Science and Technology of China (USTC), China, in 2011. From September 2011 to September 2013, he worked as a postdoc researcher in Computer Science Department at the University of Texas at San Antonio. He is currently a Professor at the EEIS Department, USTC. 

His research interests include multimedia information retrieval, computer vision, and computer game. In those fields, he has published over 100 papers in IEEE/ACM Transactions and CCF Tier-A International Conferences. He is the winner of National Science Funds of China (NSFC) for Excellent Young Scientists. He is the recepient of the Best Paper Award for ICIMCS 2012. He received the award for the Excellent Ph.D Supervisor of Chinese Society of Image and Graphics (CSIG) in 2021, and the award for the Excellent Ph.D Supervisor of Chinese Academy of Sciences (CAS) in 2022. He won the First Class Wu-Wenjun Award for Progress in Artificial Intelligence Technology in 2021. He served as the publication chair of IEEE ICME 2021 and won 2021 ICME Outstanding Service Award. He is currently an Associate Editor and a Lead Guest Editor of IEEE Transactions on Multimedia. 
\end{IEEEbiography}

\begin{IEEEbiography}[{\includegraphics[width=1in,height=1.25in,clip,keepaspectratio]{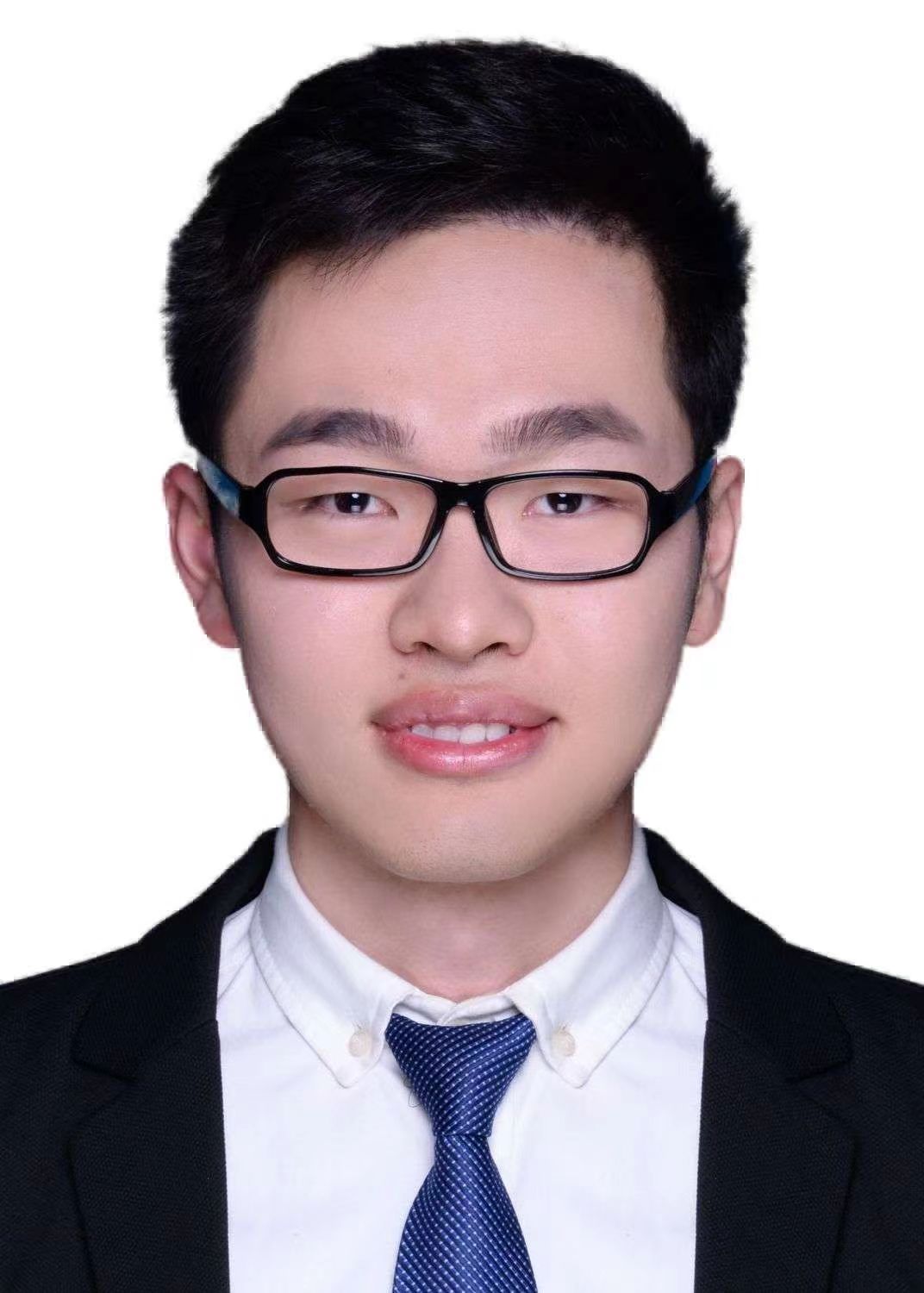}}]{Hezhen Hu} is a postdoctoral researcher at The University of Texas at Austin. He received the Ph.D. degree in information and communication engineering with the Department of Electronic Engineering and Information Science, from the University of Science and Technology of China~(USTC), in 2023. His research interests include sign language understanding, self-supervised pre-training, human-centric visual understanding, and computer vision.
\end{IEEEbiography}

\begin{IEEEbiography}[{\includegraphics[width=1in,height=1.25in,clip,keepaspectratio]{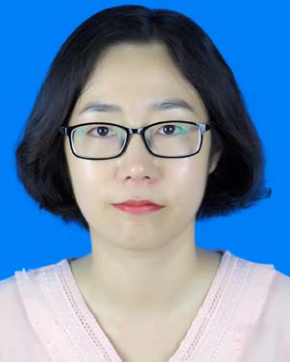}}]{Min Wang} received the BE, and PhD degrees in electronic information engineering from the University of Science and Technology of China (USTC), in 2014 and 2019, respectively. From July 2019 to 2020, she worked with Huawei Noah’s Ark Lab. She is working with the Institute of Artificial Intelligence, Hefei Comprehensive National Science Center. Her
current research interests include multimedia information retrieval and computer vision.
\end{IEEEbiography}

\begin{IEEEbiography}[{\includegraphics[width=1in,height=1.25in,clip,keepaspectratio]{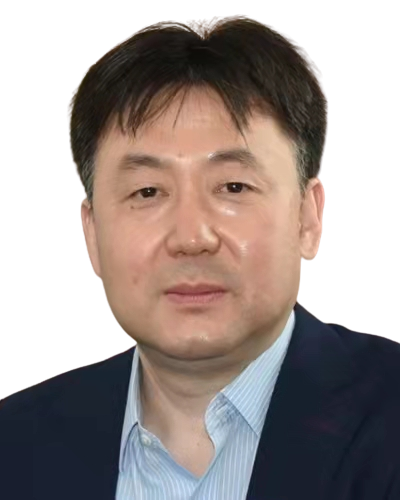}}]{Houqiang Li} (S'12, F'21) received the B.S., M.Eng., and Ph.D. degrees in electronic engineering from the University of Science and Technology of China, Hefei, China, in 1992, 1997, and 2000, respectively, where he is currently a Professor with the Department of Electronic Engineering and Information Science. 
	
His research interests include image/video coding, image/video analysis, computer vision, reinforcement learning, etc.. He has authored and co-authored over 200 papers in journals and conferences. He is the winner of National Science Funds (NSFC) for Distinguished Young Scientists, the Distinguished Professor of Changjiang Scholars Program of China, and the Leading Scientist of Ten Thousand Talent Program of China. He is the associate editor (AE) of IEEE TMM, and served as the AE of IEEE TCSVT from 2010 to 2013. He served as the General Co-Chair of ICME 2021 and the TPC Co-Chair of VCIP 2010. He received the second class award of China National Award for Technological Invention in 2019, the second class award of China National Award for Natural Sciences in 2015, and the first class prize of Science and Technology Award of Anhui Province in 2012. He received the award for the Excellent Ph.D Supervisor of Chinese Academy of Sciences (CAS) for four times from 2013 to 2016. He was the recipient of the Best Paper Award for VCIP 2012, the recipient of the Best Paper Award for ICIMCS 2012, and the recipient of the Best Paper Award for ACM MUM in 2011.
\end{IEEEbiography}

\end{document}